\documentclass[10pt,journal,compsoc]{IEEEtran}%
\usepackage[nocompress]{cite}
\usepackage{cite}
\usepackage{graphicx}
\usepackage{amsmath}
\usepackage{amssymb}
\usepackage{subfigure}
\usepackage{amsfonts}
\usepackage{placeins}
\usepackage{algorithm}
\usepackage{algpseudocode}
\usepackage{epstopdf}
\usepackage{comment}
\usepackage{amssymb,amsmath,amsthm}
\usepackage{xcolor}
\usepackage[nocompress]{cite}
\usepackage{cite}%
\providecommand{\U}[1]{\protect\rule{.1in}{.1in}}
\ifCLASSOPTIONcompsoc
\else
\fi
\ifCLASSINFOpdf
\else
\fi
\graphicspath{{figures/}}
\begin{document}

\title{Joint detection and matching of feature points in multimodal images}
\author{Elad Ben Baruch,Yosi Keller$^{\ast}$
\IEEEcompsocitemizethanks{\IEEEcompsocthanksitem E. Ben Baruch was with the Faculty
of Engineering, Ben-Gurion University of the Negev, Israel.\protect \and
\IEEEcompsocthanksitem Y. Keller is with the Faculty of Engineering, Bar Ilan University, Ramat-Gan, Israel.\protect \and
E-mail: yosi.keller@gmail.com}\thanks{Manuscript received April 19, 2005;
revised August 26, 2015.}}
\maketitle

\begin{abstract}
In this work, we propose a novel Convolutional Neural Network (CNN)
architecture for the joint detection and matching of feature points in images
acquired by different sensors using a single forward pass. The resulting
feature detector is tightly coupled with the feature descriptor, in contrast
to classical approaches (SIFT, etc.), where the detection phase precedes and
differs from computing the descriptor. Our approach utilizes two CNN
subnetworks, the first being a Siamese CNN and the second, consisting of dual
non-weight-sharing CNNs. This allows simultaneous processing and fusion of the
joint and disjoint cues in the multimodal image patches. The proposed approach
is experimentally shown to outperform contemporary state-of-the-art schemes
when applied to multiple datasets of multimodal images. It is also shown to
provide repeatable feature points detections across multi-sensor images,
outperforming state-of-the-art detectors. To the best of our knowledge, it is
the first unified approach for the detection and matching of such images.

\end{abstract}



\IEEEdisplaynontitleabstractindextext
\IEEEpeerreviewmaketitle

\section{Introduction}

\label{sec:introduction}

\begin{figure}[t]
\centering\includegraphics[width=0.5\textwidth]{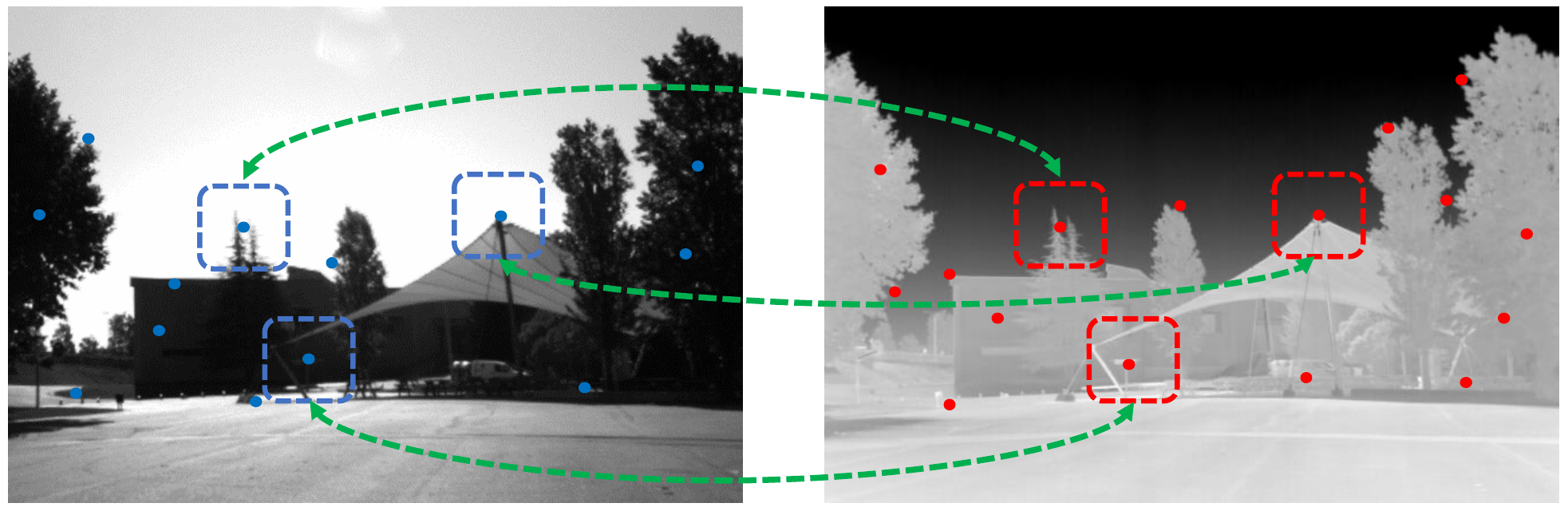}\caption{\textbf{The
multisensor patch matching problem.} The matched optical (left) and IR (right)
images differ by significant appearance changes due to the dissimilar physical
characteristics captured by the different sensors. The images are part of the
LWIR-RGB dataset \cite{aguilera2012multispectral}. The feature points in both
images were detected using the proposed scheme.}%
\label{fig:visVsIrDemo}%
\end{figure}

The detection and matching of feature points in images is a fundamental task
in computer vision and image processing that is applied in common computer
vision tasks such as image registration \cite{zitova2003image}, dense image
matching, \cite{vaquero2010survey} and 3D reconstruction
\cite{agarwal2011building}, to name a few. The term \textit{feature point
}relates to the center of an image patch, that is expected to be salient and
repeatedly detected in multiple images of the same scene, which might differ
by pose and appearance \cite{lowe2004distinctive}. A \textit{detector}
identifies the spatial location of a feature point, and the surrounding patch
is encoded by a \textit{descriptor}. The detection and matching of feature
points in multi-modal images, as depicted in Fig. \ref{fig:visVsIrDemo}, is of
particular interest in remote sensing
\cite{irani1998robust,li1995contour,keller2006multisensor,hasan2012modified}
and medical imaging \cite{sotiras2013deformable}, as the fusion of such
images, provides information synergy. The acquisition of the same scenes by
different sensors might result in significant appearance variations, that are
often nonlinear and unknown apriori, such as non-monotonic intensity mappings,
contrast reversal, and non-corresponding edges and textures.

The registration of the multimodal input images $\mathbf{I}_{\mathbf{1}}$ and
$\mathbf{I}_{\mathbf{2}}$ can be formulated as the estimation of a parametric
(rigid, affine, etc.) global transformation $T$, by minimizing an appearance
invariant similarity measures $\phi(\mathbf{I}_{\mathbf{1}},\mathbf{I}%
_{\mathbf{2}})$%
\begin{equation}
T^{\ast}=\arg\min_{T}\phi(\mathbf{I}_{\mathbf{1}},T(\mathbf{I}_{\mathbf{2}})).
\label{eq:GlobalAllignment}%
\end{equation}
$T(\mathbf{I}_{\mathbf{2}})$ is the image $\mathbf{I}_{\mathbf{2}}$ warped
towards $\mathbf{I}_{\mathbf{1}}$, according to $T$. Gradient-based approaches
were applied by Irani et al. \cite{irani1998robust}, and Keller et al.
\cite{keller2006multisensor} to appearance-invariant image representations to
solve Eq. \ref{eq:GlobalAllignment} iteratively.

Other multimodal registration schemes are based on matching local image
features such as patches, contours \cite{li1995contour}, and corners. Such
approaches match the sets of interest points $S_{1}\in\mathbf{I}_{\mathbf{1}}$
and $S_{2}\in\mathbf{I}_{\mathbf{2}}$, where each feature point is first
detected and then encoded by a robust appearance-invariant descriptor. A pair
of descriptors can be matched by computing their $L_{2}$ distance.
Such descriptors were commonly derived by extending unimodal
descriptors such as SIFT \cite{lowe2004distinctive} and Daisy
\cite{tola2010daisy} to the multimodal case
\cite{chen2009real,hossain2011improved,aguilera2012multispectral,lghd2015,kwon2016dude}%
. \

Convolutional Neural Networks (CNNs) were applied to feature point matching
\cite{Aguilera_cvprw_2016}, by training data-driven multimodal image
descriptors. These CNNs are trained by optimizing a Hinge Loss applied to an
$L_{1}$ or $L_{2}$ metrics, while others
\cite{zagoruyko2015learning,aguilera2017cross,en2018ts,AFD-Net} aim to compute
a similarity score between image patches by optimizing the Cross-Entropy loss
by classifying the pairs of patches as same/not-same. Such approaches utilize
Siamese CNNs \cite{Aguilera_cvprw_2016} consisting of weight sharing sub-networks.

The upside of $L_{2}$-based representations compared to those computed using
the Cross-Entropy loss is their reduced computational complexity when applied
to match sets of feature points detected in a pair or set of images. As an
image typically contains $n=300-500$ feature points, matching a pair of images
requires $n^{2}/2$ point-to-point similarity evaluations. $K$ nearest
neighbors (KNN) similarity search via $L_{2}$-based representations can be
computationally accelerated using metric embedding schemes such as Locality
Sensitive Hashing (LSH) \cite{LSH} and MinHash \cite{minhash}.

Feature detectors are commonly applied to each image separately, without
relating the detections in one modality to the other. A detector aims to
detect the spatial location of the feature point and estimate its local scale
and orientation. The location is often determined using a corner detector
\cite{lowe2004distinctive}, and the local scale is computed using the
Difference of Gaussians (DoG) operator and its approximations
\cite{Lindeberg1993,lowe2004distinctive}.

In this work, we propose a CNN-based metric learning approach for the joint
detection and matching of feature points in multimodal images, using \textit{a
single forward pass}. When applied to image patches, the proposed scheme
computes the corresponding descriptors and similarity. But, when applied to
full-scale images, the proposed fully convolutional CNN computes a grid of
descriptors. By propagating back through the corresponding CNN activations and
layers, the locations of feature points corresponding to each descriptor are
detected. As the CNN is trained to optimize the descriptors' matching, the
proposed detector is tightly coupled with the feature descriptor, in contrast
to classical approaches such as SIFT \cite{lowe2004distinctive} and its (many)
extensions, where the detector is computed separately before computing the
descriptor. To the best of our knowledge, we introduce the first unified
approach for the detection and matching of multi-modality images.

In particular, we present the Hybrid CNN architecture consisting of both a
Siamese\textit{\ }sub-network and a dual-channel\textit{\ }non-weight-sharing
asymmetric sub-network. The use of the asymmetric sub-network is due to the
inherent asymmetry in the multisensor matching problem, where the
heterogeneous inputs might differ significantly, and thus require different
processing implemented by the asymmetric sub-network. In particular, each
branch of the asymmetric sub-network estimates a \textit{modality-specific}
adaptive representation of the multisensor patches.

Thus, we aim to leverage both the joint and disjoint attributes in the
multimodal images, using the Siamese and Asymmetric subnets, respectively.
Siamese sub-networks were previously shown \cite{Aguilera_cvprw_2016} to yield
accurate matching results, and are outperformed by the proposed Hybrid scheme.
The Siamese and Asymmetric subnets are trained by corresponding losses, and
their outputs are merged and optimized to yield a fused image representation.

In particular, we propose the following contributions:

\textbf{First, }we present a novel approach for the joint detection and
matching of feature points in multi-modality images.

\textbf{Second}, the proposed scheme is implemented using a novel Hybrid CNN
architecture consisting of both a Siamese and asymmetric sub-networks, able to
leverage both the joint and disjoint cues in multimodal patches, to determine
their similarity.

\textbf{Third, }we show that training the proposed Hybrid CNN by multi-loss
learning improves the descriptors' matching accuracy.

\textbf{Last, }the proposed scheme was experimentally shown to outperform
contemporary approaches when applied to state-of-the-art multimodal image
matching benchmarks \cite{Aguilera_cvprw_2016,aguilera2017cross,en2018ts} and
feature points detection schemes. The corresponding source code was made
publicly available\footnote{https://github.com/eladbb/HybridSiamese}.

\section{Related work}

\label{sec:Previous and Related work}

\subsection{Appearance-invariant image representations}

The matching of multi sensor images has been studied in a gamut of works.
Earlier, unsupervised approaches were mostly based \ on deriving
appearance-invariant image representations of multisensor images, that
utilized salient image edges and contours. Thus, Irani et
al.~\cite{irani1998robust} suggested a coarse-to-fine scheme for estimating
the global parametric motion (affine, rigid) between multimodal images, using
the magnitudes of directional derivatives as a robust image representation.
The correlation between these representations is maximized using iterative
gradient methods and a coarse-to-fine formulation.

The \textquotedblleft Implicit Similarity\textquotedblright\ formulation by
Keller et al.~\cite{keller2006multisensor} is an iterative scheme utilizing
gradient information for global alignment. A set of pixels with maximal
gradient magnitude is detected in one of the input images, rather than
contours and edges as in \cite{irani1998robust}. The gradient of the
corresponding points in the second image is maximized with respect to a global
parametric motion, without explicitly maximizing a similarity measure. The
seminal work of Viola and Wells \cite{Viola:1997:AMM:263008.263015} proposed
an appearance-robust image representation based on the statistical
representations of the images. The mutual information between these
representations was optimized with respect to the motion parameters.

Modality-invariant local descriptors were derived by modifying the
SIFT~descriptor \cite{chen2009real,MI-SIFT}. Contrast-invariance was achieved
by mapping the gradient orientations of the interest points from $[0,2\pi)$ to
$[0,\pi)$. Hasan et al. showed that such descriptors mitigate the matching
accuracy \cite{hossain2011improved}, and further modified the SIFT descriptor
\cite{hasan2012modified} by thresholding gradient values to reduce the effect
of strong edges. An enlarged spatial window with additional sub-windows was
used to improve the spatial resolution. Aguilera et
al.~\cite{aguilera2012multispectral} used a histogram of contours and edges
instead of a histogram of gradients to avoid the ambiguity of the SIFT
descriptor when applied to multimodal images, while the dominant orientation
was determined similarly. This approach was extended by the same authors by
using multi-oriented and multi-scale Log-Gabor filters \cite{lghd2015}.

\subsection{Structural similarity}

Geometrical structure was used as an appearance-invariant representation,
where the similarity was quantified by structural similarity. Hence, the Self
Similarity Matching (SSM) approach by Shechtman et
al.~\cite{shechtman2007matching} is a geometric descriptor encoding the local
geometric structure around a feature point in an image, by correlating a
central patch to all adjacent patches within a predefined radius. The Dense
Adaptive Self-Correlation (DASC) descriptor, by Kim et al.~\cite{kim2015dasc},
extended the SSM approach\textbf{, }by computing the self-similarity measure
as an adaptive self-correlation between randomly sampled patches.

\subsection{Deep learning approaches}

With the emergence of CNNs as the state-of-the-art approach to a gamut of
computer vision problems, CNNs were applied to patch matching. Siamese CNNs
such as HardNet {\ \cite{hardnet}, }L2-Net {\ \cite{8100132} and }MatchNet
\cite{MatchNet} were applied to matching single modality images, significantly
outperforming the classical handcrafted image features.

Zagoruyko and Komodakis \cite{zagoruyko2015learning} proposed several CNN
architectures for multisensor feature matching, using a Siamese CNN with a
Contrastive or Cross-Entropy losses, for matching single modality patches, and
a CNN, where the input patches are stacked as different image channels.
Aguilera et al.~\cite{Aguilera_cvprw_2016} applied the approaches by Zagoruyko
and Komodakis \cite{zagoruyko2015learning} to matching multimodal patches and
showed that the resulting CNN outperformed the state-of-the-art multimodal descriptors.

To alleviate the computational complexity of the stacked approach when applied
to sets of feature points, Aguilera et al. proposed the Q-Net CNN
\cite{aguilera2017cross} that was trained using an $L_{2}$ loss. The Q-Net CNN
consists of four weight sharing sub-networks and two corresponding pairs of
input patches, that allow hard negative mining \cite{hardnet}. This approach
was shown to achieve state-of-the-art accuracy when applied to the Vis-Nir
benchmark \cite{Aguilera_cvprw_2016}.

En et al. \cite{en2018ts} introduced a simplified Hybrid-like CNN, denoted as
TS-Net, for multimodal patch matching, utilizing a Cross-Entropy loss.
Contrary to the proposed scheme, this approach does not compute $L_{2}%
$-optimized patch encodings, that are essential for matching images, typically
consisting of 300-500 feature points. Each of the sub-networks outputs a
scalar Softmax prediction that are merged using a FC layer. In contrast, we
utilize a different architecture where feature maps are fused, and apply
auxiliary losses (that are not used by En et al. \cite{en2018ts}).

Metric learning was applied by Quan et al. \cite{SCFDM} to learn the shared
feature space of multi-spectral patches, by progressively comparing spatially
connected features, using a discrimination constraint (SCFDM). This approach
was extended by the same authors, by deriving the AFD-Net \cite{AFD-Net}, that
learns multiscale joint multi-spectral features using a CNN consisting of two
subnetworks. The activations maps at different layers are subtracted, and the
differences are propagated through multiple FC layers. Thus, this approach
does not compute a descriptor, and the matching of two images having $n$
feature points each, entails $n^{2}/2$ forward passes of the CNN, in contrast,
to the single forward pass required by the proposed scheme. A novel metric
learning cost function denoted the Exponential Loss, was introduced by Wang et
al. for patch matching \cite{Wang_2019_ICCV}. Using a corresponding hard
negative mining scheme, it was shown to provide SOTA accuracy on the VIS-NIR
dataset \cite{Aguilera_cvprw_2016}.

The joint detection and matching of feature points in \textit{single} modality
images was first suggested by Yi et al. \cite{LIFT} that proposed the LIFT
scheme using a Siamese CNN. LIFT consists of three successive parts: a feature
point detector, followed by an estimate of the detected point's orientation
(similar to the \textit{dominant orientation} in SIFT), and the feature point
descriptor. Dusmanu et al. \cite{D2-Net} also proposed a Siamese network,
where the descriptors are trained using a triplet ranking loss, and the
features are detected as the local maxima of the last activation map. The
corresponding CNN was implemented without pooling layers to relate the
detections in the last activation map to the source (finest) image resolution.
The model was trained using pixel correspondences computed by large-scale SfM
reconstructions. An image pyramid consisting of three resolutions is used to
account for scale variations, by computing the descriptors in all scales.

Simeoni et al. \cite{Simeoni} proposed a multi-scale feature detection scheme
for \textit{single} modality images by detecting local maxima in the
activation maps over multiple activation layers. The activations are localized
per channel using the maximally stable extremal regions (MSER) blob detector.
As in \cite{D2-Net}, a Siamese network is used to match the corresponding
detections in the training images.

\section{Detection and matching of multi-modal feature points}

\label{sec:Proposed Model}

\begin{figure*}[th]
\centering\includegraphics[width=0.7\textwidth]{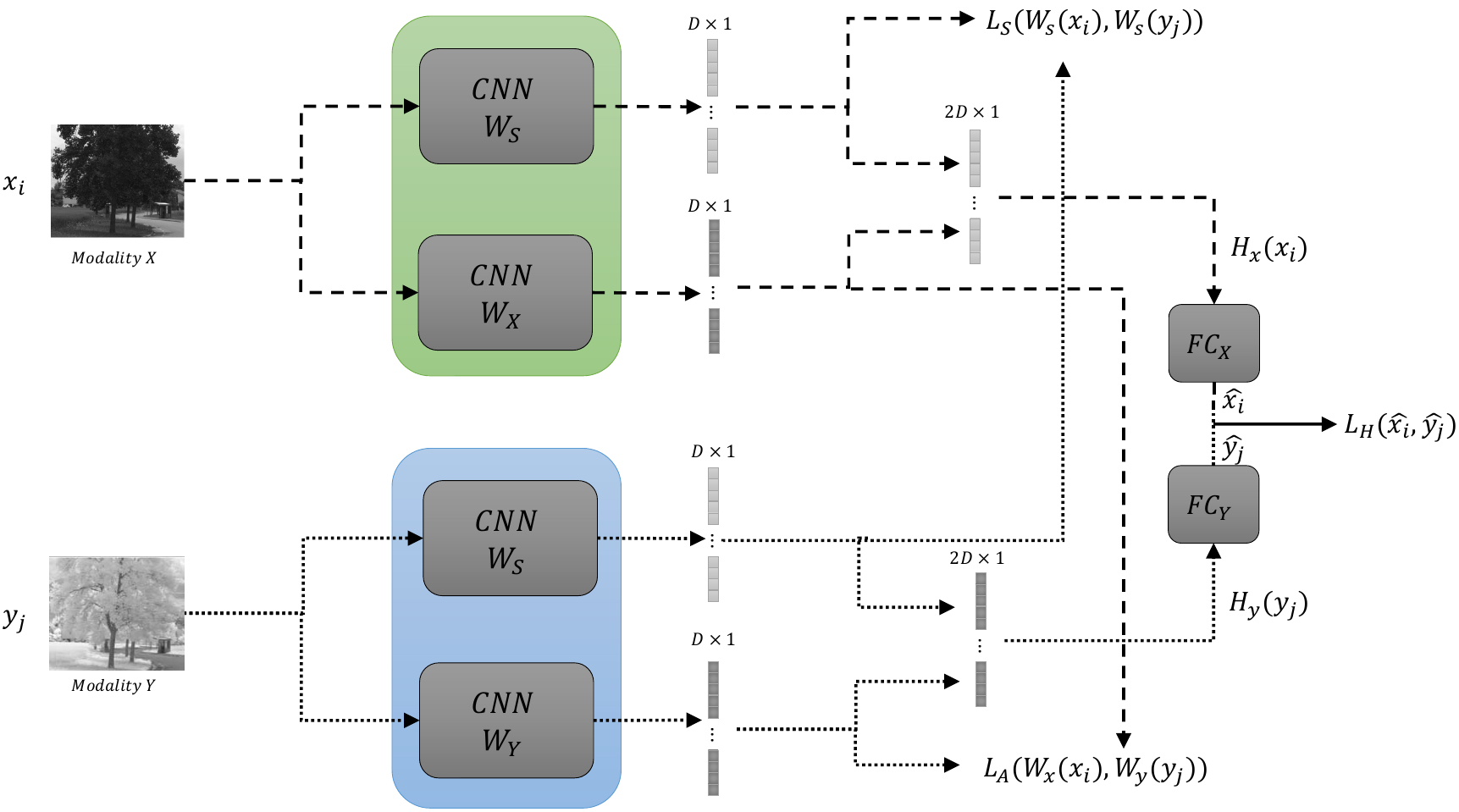}\caption{\textbf{The
proposed hybrid matching model.} The model consists of two sub-networks: a
siamese subnetwork and an asymmetric subnetwork with non-shared weights. The
siamese branch consists of a pair of $W_{s}$ CNNs and is trained by the loss
$L_{s}$. The asymmetric branch consists of the $W_{x}$ and $W_{y}$ CNNs and is
trained by the loss $L_{A}$. The symmetric and asymmetric representations are
merged and trained by the loss $L_{H}$.}%
\label{fig:hybridModelIllustration}%
\end{figure*}

Let $\mathbf{x}_{i}\in\mathbf{I}_{\mathbf{1}}$ and $\mathbf{y}_{i}%
\in\mathbf{I}_{\mathbf{2}}$ be a pair of multi-dimensional image patches
acquired by different modalities. We aim to compute a corresponding
representation, $\widehat{\mathbf{x}_{i}}$ and $\widehat{\mathbf{y}_{i}}$,
respectively. The Hybrid network learns both the joint and disjoint
characteristics of the multisensor patches, using both Siamese (symmetric) and
asymmetric (non-weight-sharing) networks. The Siamese network learns
\textit{the same} mapping for both input modalities, denoted as ${W}_{{s}%
}(\mathbf{x}_{i})$ and ${W}_{{s}}(\mathbf{y}_{i})$ in Fig.
\ref{fig:hybridModelIllustration}, respectively, allowing to encode the
\textit{same} characteristics between the images. The asymmetric network
estimates \textit{different}, modality-specific representations, ${W}%
_{x}(\mathbf{x}_{i})$ and ${W}_{y}(\mathbf{y}_{i})$, for each modality,
respectively. The outputs of the symmetric and asymmetric sub-networks are
concatenated
\begin{align}
H_{x}(\mathbf{x}_{i})  &  =\left[  {W}_{{s}}(\mathbf{x}_{i}),{W}%
_{x}(\mathbf{x}_{i})\right]  ^{T},\\
H_{y}(\mathbf{y}_{i})  &  =\left[  {W}_{{s}}(\mathbf{y}_{i}),{W}%
_{y}(\mathbf{y}_{i})\right]  ^{T},\nonumber
\end{align}
and the resulting Hybrid representation is given by
\begin{align}
\widehat{\mathbf{x}_{i}}  &  =FC_{x}(H_{x}(\mathbf{x}_{i})),\label{equ:hybrid}%
\\
\widehat{\mathbf{y}_{i}}  &  =FC_{y}(H_{y}(\mathbf{y}_{i}))\nonumber
\end{align}
where $FC_{x}$ and $FC_{y}$ are FC layers. 

The proposed CNN\ is fully convolutional and can thus be applied to images of
varying dimensions. When applied to $64\times64$ image patches, the Hybrid CNN
yields a pair of $\{\widehat{\mathbf{x}_{i}},\widehat{\mathbf{y}_{i}}\}\in%
\mathbb{R}
^{128}$ descriptors. But, when applied to larger images, an activation map
$\{\widehat{\mathbf{x}_{i,j}},\widehat{\mathbf{y}_{i,j}}\}\in%
\mathbb{R}
^{128}$ is computed. Each descriptor relates to a particular patch in the
image, according to its footprint\textbf{. }We show that by backtracking
through the CNN\ activations, down to the input layer, we detect the location
of the corresponding feature points in the finest image resolution, as
detailed in Section \ref{subsec:detection}.

The proposed CNN is trained using multiple losses. The losses $L_{S}$ and
$L_{A}\,$in Fig. \ref{fig:hybridModelIllustration} optimize the symmetric and
asymmetric subnetworks, respectively. This was shown to improve the matching
accuracy, as these losses ($L_{S}$ and $L_{A}$) optimize CNNs having less
number of parameters compared to the full Hybrid CNN. The unified Hybrid
representation, as in Eq. \ref{equ:hybrid}, is trained using the $L_{H}$ loss.

We used both the Binary Cross-Entropy (BCE) and the Contrastive Loss (CL)
\cite{1640964}. The CL is give by%
\begin{multline}
L_{C}\left(  x_{i},x_{j},x_{k}\right)  =\frac{1}{\left\vert Pos\right\vert }%
{\displaystyle\sum\limits_{x_{i},x_{j}\in Pos}}
D^{2}\left(  x_{i},x_{j}\right)  +\\
\frac{1}{\left\vert Neg\right\vert }%
{\displaystyle\sum\limits_{x_{i}\in Pos,x_{k}\in Neg}}
\left[  m-D^{2}\left(  x_{i},x_{k}\right)  \right]  _{+}%
\end{multline}
where $\left[  \cdot\right]  _{+}$ is the Hinge loss operator, and $D\left(
x_{i},x_{j}\right)  $ is the $L_{2}$ distance between the embeddings. $m$ is a
predefined threshold that is commonly defined as $m=1$ for $L_{2}$ normalized
embeddings, as in our CNN.

All losses $\{L_{S},L_{A},L_{H}\}$ are either all BCE or CL. The choice of the
loss relates to the particular task of the multi-modality descriptors
$\widehat{\mathbf{x}_{i}}$ and $\widehat{\mathbf{y}_{i}}$. For the BCE-based
descriptors, a Softmax layer outputs the matching probability of the input
patches, while the CL yields a descriptor embedded in a Euclidean space. Such
descriptors can be utilized in efficient large-scale descriptor retrieval
schemes, where $K$-nearest-neighbors (KNN) $L_{2}$ search can be efficiently
implemented using LSH \cite{LSH} and MinHash \cite{minhash}.

\subsection{CNN architecture}

\label{subsec:CNN architecture}

The proposed networks consist of a Siamese (symmetric) and asymmetric
networks, as depicted in Fig. \ref{fig:hybridModelIllustration}. The Siamese
network consists of the two weight-sharing networks $W_{s}$, while the
asymmetric network consists of the non-weight-sharing networks $\{W_{X}%
,W_{Y}\}$ applied to the inputs $X$ and $Y$, respectively. When the losses
$\{L_{S},L_{A},L_{H}\}$ are all Contrastive losses, $W_{X},W_{Y}$ and $W_{s}$
are detailed in Table \ref{tab:L2BranchArchitecture}. \begin{table}[t]
\centering
\par%
\begin{tabular}
[c]{|c|c|c|c|c|}\hline
\textbf{Layer} & \textbf{Output} & \textbf{Kernel} & \textbf{Stride} &
\textbf{Pad}\\\hline\hline
\multicolumn{1}{|l|}{Conv0} & $64\times64\times32$ & $5\times5$ & 1 &
2\\\hline
\multicolumn{1}{|l|}{Pooling} & $32\times32\times32$ & $3\times3$ & 2 &
-\\\hline
\multicolumn{1}{|l|}{Conv1} & $32\times32\times64$ & $5\times5$ & 1 &
2\\\hline
\multicolumn{1}{|l|}{Pooling} & $16\times16\times64$ & $3\times3$ & 2 &
-\\\hline
\multicolumn{1}{|l|}{Conv2} & $16\times16\times128$ & $3\times3$ & 1 &
1\\\hline
\multicolumn{1}{|l|}{Pooling} & $8\times8\times128$ & $3\times3$ & 2 &
-\\\hline
\multicolumn{1}{|l|}{Conv3} & $6\times6\times256$ & $3\times3$ & 1 & 0\\\hline
\multicolumn{1}{|l|}{Conv4} & $4\times4\times256$ & $3\times3$ & 1 & 0\\\hline
\multicolumn{1}{|l|}{FC} & $1\times1\times128$ & - & - & -\\\hline
\multicolumn{1}{|l|}{Unit norm} & $1\times1\times128$ & - & - & -\\\hline
\end{tabular}
\caption{\textbf{The CNN architecture of the sub-networks using the
Contrastive Loss.} Each sub-network accepts a $64\times64$ patch and outputs a
$128\times1$ descriptor. ReLU activations are applied after each pooling
layer, as well as after the Conv3 layer.}%
\label{tab:L2BranchArchitecture}%
\end{table}Similarly, when the losses $\{L_{S},L_{A},L_{H}\}$ are all BCE
losses, $W_{X},W_{Y}$ and $W_{s}$ are given by Table
\ref{tab:SoftmaxBranchArchitecture}, and the overall training loss is given by%
\begin{equation}
L=L_{S}+L_{A}+L_{H}%
\end{equation}
\begin{table}[t]
\centering
\par%
\begin{tabular}
[c]{|c|c|c|c|c|}\hline
\textbf{Layer} & \textbf{Output} & \textbf{Kernel} & \textbf{Stride} &
\textbf{Pad}\\\hline\hline
\multicolumn{1}{|l|}{Conv0} & $64\times64\times32$ & $5\times5$ & 1 &
2\\\hline
\multicolumn{1}{|l|}{Pooling} & $32\times32\times32$ & $3\times3$ & 2 &
-\\\hline
\multicolumn{1}{|l|}{Conv1} & $32\times32\times64$ & $5\times5$ & 1 &
2\\\hline
\multicolumn{1}{|l|}{Pooling} & $16\times16\times64$ & $3\times3$ & 2 &
-\\\hline
\multicolumn{1}{|l|}{Conv2} & $16\times16\times128$ & $3\times3$ & 1 &
1\\\hline
\multicolumn{1}{|l|}{Pooling} & $8\times8\times128$ & $3\times3$ & 2 &
-\\\hline
\multicolumn{1}{|l|}{Conv3} & $6\times6\times256$ & $3\times3$ & 1 & 0\\\hline
\multicolumn{1}{|l|}{Conv4} & $4\times4\times256$ & $3\times3$ & 1 & 0\\\hline
\multicolumn{1}{|l|}{Conv5} & $2\times2\times256$ & $3\times3$ & 1 & 0\\\hline
\multicolumn{1}{|l|}{FC} & $1\times1\times128$ & - & - & -\\\hline
\end{tabular}
\caption{\textbf{The CNN architecture of the sub-networks using the
Cross-Entropy loss.} Each sub-network accepts a $64\times64$ patch and outputs
a $128\times1$ descriptor. ReLU activations are applied after each pooling
layer, as well as after the Conv3 and Conv4 layers.}%
\label{tab:SoftmaxBranchArchitecture}%
\end{table}

\subsection{The detection of feature points in multi-modality images}

\label{subsec:detection}

A feature point $\mathbf{v}=\left(  l,k\right)  $ is a pixel location, such
that the feature points $\mathbf{v}_{x}\in\mathbf{I}_{\mathbf{1}}$ and
$\mathbf{v}_{y}\in\mathbf{I}_{\mathbf{2}}$ relate to the descriptors
$\widehat{\mathbf{x}}$ and $\widehat{\mathbf{y}}$, respectively. The
fundamental property of a feature point is its \textit{repeatability}
\cite{lowe2004distinctive,1498756}, implying that corresponding points
$\mathbf{v}_{x}\in\mathbf{I}_{\mathbf{1}}$ and $\mathbf{v}_{y}\in
\mathbf{I}_{\mathbf{2}}$ relate to joint image content in both images.

We propose to detect the feature points by analyzing the joint representation
encoded by the Siamese subnetworks $\widehat{\mathbf{x}_{i}^{s}}={W}_{{s}%
}(\mathbf{x}_{i})$ and $\widehat{\mathbf{y}_{i}^{s}}={W}_{{s}}(\mathbf{y}%
_{i})$ (as in Fig. \ref{fig:hybridModelIllustration}) to detect $\mathbf{v}%
_{x}\in\mathbf{I}_{\mathbf{1}}$, and $\mathbf{v}_{y}\in\mathbf{I}_{\mathbf{2}%
}$, respectively. We also applied the asymmetric subnetworks ${W}_{{x}%
}(\mathbf{x}_{i})$ and ${W}_{{y}}(\mathbf{y}_{i}),$ resulting in less accurate results.

The proposed Hybrid patch-matching scheme is implemented by \textit{fully
convolutional} CNNs. Thus, when ${W}_{{s}}(\mathbf{x}_{i})$ and ${W}_{{s}%
}(\mathbf{y}_{i})$ are applied to $64\times64$ patches, the result is $\in%
\mathbb{R}
^{128\times1}$. But, when applied to images $\in%
\mathbb{R}
^{W\times H\times3},$ their outputs are activation maps $\in%
\mathbb{R}
^{\frac{W}{64}\times\frac{H}{64}\times C}$, where $C$ is the number of
channels in the activation map. Hence, each point in ${W}_{{s}}(\mathbf{x}%
_{i})$ and ${W}_{{s}}(\mathbf{y}_{i})$ encodes an image patch $\mathbf{P}\in%
\mathbb{R}
^{64\times64\times3}$.\ As the Hybrid CNN is trained to optimize the matching
between the $\mathbf{x}$ and $\mathbf{y}$ modalities, ${W}_{{s}}$ is also
jointly optimized to extract the most informative features from the
corresponding images patches $\left\{  \mathbf{P}_{i,j}\right\}  $. The
locations of the largest activations are the more informative, and the
locations with smaller values are pruned by the max-pooling layers.

Thus, the core of our approach is to detect the informative feature points in
the input image given the activation map $\widehat{\mathbf{x}_{i}^{s}}%
={W}_{{s}}(\mathbf{x}_{i})\in%
\mathbb{R}
^{\frac{W}{64}\times\frac{H}{64}\times C}$ . \ The detection is applied by
backtracking the spatial locations of the activation $\widehat{\mathbf{x}%
_{i}^{s}}$ through ${W}_{{s}},$ down to the first activation layer to identify
the pixels contributing to ${W}_{{s}}(\mathbf{x}_{i})$. In contrast to those
that were pruned by the max-pooling layers.

\subsubsection{Backtracking for feature detection}

\label{subsubsec:Backtracking}

The ${W}_{{s}}$ CNN, detailed in Tables \ref{tab:L2BranchArchitecture} and
\ref{tab:SoftmaxBranchArchitecture}, consists of padded convolutions and
pooling layers. Let $\widehat{\mathbf{x}}^{l}\in%
\mathbb{R}
^{w_{l}\times h_{l}\times c_{l}}$\thinspace\ be the activation map at level
$l$ of spatial dimensions $w_{l}\times h_{l}$ and $c_{l}$ channels. Spatially
symmetric (i.e. $3\times3$) padded convolutions do not change the locations of
the activations, such that each element $\widehat{\mathbf{x}}_{i,j}^{l}%
\in\widehat{\mathbf{x}}^{l}$ corresponds to the same spatial location in the
preceding activation map $\widehat{\mathbf{x}}_{i,j}^{l-1}\in\widehat
{\mathbf{x}}^{l-1}$. This implies that when backtracking through a symmetric
convolution layer, we have that $\widehat{\mathbf{x}}_{i,j}^{l}\rightarrow
\widehat{\mathbf{x}}_{i,j}^{l-1}$.\begin{figure}[th]
\centering\includegraphics[width=0.4\textwidth]{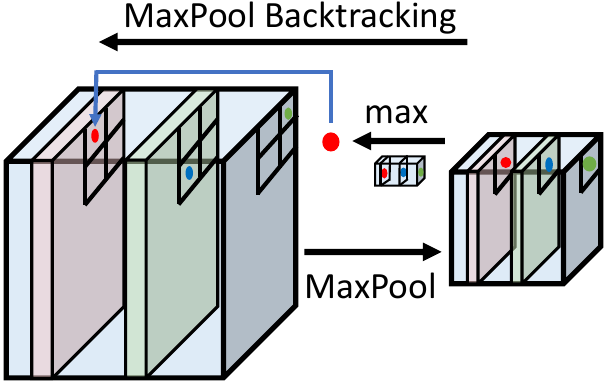}\caption{\textbf{Backtracking
a }$2\times2$\textbf{ max-pooling layer.} In the forward pass (left to right),
multiple spatial locations are mapped by the max-pooling layer to a single
spatial location. In the backtracking pass (right to left), the maximum over
all channels in a particular spatial location (red dot), is the one propagated
back.}%
\label{fig:maxpool}%
\end{figure}

In contrast, max-pooling layers propagate the content of \textit{multiple}
spatial locations in the activation map $\widehat{\mathbf{x}}^{l-1}$ to a
\textit{single} spatial location in the succeeding layer $\widehat{\mathbf{x}%
}^{l}$, as depicted in Fig. \ref{fig:maxpool}. Let $\widehat{\mathbf{x}}%
^{l-1}\in%
\mathbb{R}
^{W\times H\times C}$ and $\widehat{\mathbf{x}}^{l}\in%
\mathbb{R}
^{\frac{W}{k}\times\frac{H}{k}\times C}$ be the input and output layers,
respectively, of a $k\times k$ max-pooling layer. For each $k\times
k\times1\in$ $\widehat{\mathbf{x}}^{l-1}$, the entry having the maximal value
is propagated forward. Thus, the entries in a \textit{single} particular
spatial location $\widehat{\mathbf{x}}^{l}\left(  w,h,c\right)  ,$ $c=1..C$
might correspond to \textit{multiple} spatial locations in $\widehat
{\mathbf{x}}^{l-1}$. To relate these entries to a \textit{single} spatial
location in $\widehat{\mathbf{x}}^{l-1}$, we only backtrack the spatial
location of the entry $\widehat{\mathbf{x}}^{l}\left(  \frac{w}{k},\frac{h}%
{k},c^{\ast}\right)  $ having the maximal activation
\begin{equation}
c^{\ast}=\arg\underset{c}{\max}\left\{  \widehat{\mathbf{x}}^{l}\left(
\frac{w}{k},\frac{h}{k},c\right)  \right\}  ,c=1..C,
\end{equation}
corresponding to a \textit{single} spatial location in $\widehat{\mathbf{x}%
}^{l-1}$, as in Fig. \ref{fig:maxpool}.

The feature points are given by the locations of the backtracked maximal
activations $\mathbf{V}\in%
\mathbb{R}
^{\frac{W}{64}\times\frac{H}{64}\times2}$. The corresponding activations are
the \textit{detections scores,} similar to the DoG detections scores used in
the SIFT DoG-based detector \cite{lowe2004distinctive}. It is common to only
utilize a subset of the `best' detected points, by choosing the $N$ feature
points having the largest detection score, or the points having a detection
score larger than a predefined threshold.

The proposed detection approach differs from Dusmanu et al. \cite{D2-Net} that
compute local (per layer) detection scores, and merge the scores from
different layers, having different spatial resolutions, by bilinearly
interpolating the lower-resolution score maps. Simeoni et al. \cite{Simeoni}
utilize a CNN\ without pooling layers to avoid backtracking through the
activation layers. Both these schemes are only applicable to single-modality images.

\section{Experimental Results}

\label{sec:Experiments}

The proposed Hybrid scheme was experimentally verified by applying it to
multi-spectral image datasets and benchmarks used in contemporary
state-of-the-art matching schemes. The first was suggested by Aguilera et al.
\cite{Aguilera_cvprw_2016} consisting of a set of matching and non-matching
pairs of patches, extracted from nine categories of the public VIS-NIR scene
dataset \cite{brown2011multi}. The feature points were detected by an interest
point detector and matched manually. We also used the Vehicle Detection in
Aerial Imagery (VEDAI) \cite{razakarivony2016vehicle} dataset of multispectral
aerial images and the CUHK \cite{wang2009face} dataset consisting of 188 faces
and corresponding sketches drawn by artists. These multimodal datasets are
spatially pre-aligned, same as the VIS-NIR dataset, and were used by En et al.
\cite{en2018ts} to create annotated training and test sets by extracting
corresponding pairs of patches on a uniform grid. We denote the
uniformly-extracted VIS-NIR dataset by En et al. as Uniform VIS-NIR (UVIS-NIR)
to differentiate it from the one by Aguilera et al. \cite{Aguilera_cvprw_2016}.

We evaluated the matching of the proposed scheme following the experimental
setups and datasets used by Aguilera et al.
\cite{Aguilera_cvprw_2016,aguilera2017cross} and En et al. \cite{en2018ts} and
the results are detailed in Sections \ref{sec:Vis-Nir benchmark} and
\ref{sec:ani}, respectively. The matching quality is quantified by the false
positive rate at 95\% recall (FPR95), same as in
\cite{Aguilera_cvprw_2016,aguilera2017cross}. FPR95 evaluate how well our
approach distinguishes correct and incorrect matches. The source code of the
proposed scheme was made publicly
available\footnote{https://github.com/eladbb/HybridSiamese}.

We compare the results of the proposed Hybrid scheme using both Contrastive
and Cross-Entropy losses to contemporary state-of-the-art approaches in
Sections \ref{sec:Vis-Nir benchmark} and \ref{sec:ani}. The detection accuracy
and an ablation study are discussed in Sections
\ref{subsec:Features detection} and \ref{sec:Auxiliary}.

\subsection{Training}

\label{subsec:training}

The Hybrid CNN was trained using stochastic gradient descent with a momentum
of 0.9, batch size of 128, learning rate of $0.01$ and weight decay of 0.0005,
where the same hyperparameters were used for training both the Contrastive and
Cross-Entropy losses. The Hybrid model was trained for 40 and 100 epochs, for
the Contrastive and Cross-Entropy losses, respectively. In both setups,
patches of 64x64 pixels were cropped and augmented by joint random rotations
of $90%
{{}^\circ}%
$, as well as horizontal and vertical flipping. The patches of each imaging
modality were normalized separately by subtracting their mean. In each
training set, as in Sections \ref{sec:Vis-Nir benchmark} and \ref{sec:ani},
the positive pairs of patches are given, and any other pairing can be
considered negative.

We start the training using random negative pairs, the same number as the
positive ones. A training batch consists of pairs of randomly chosen
corresponding (positive) pairs of patches, and negative pairs are chosen by
randomly ordering the patches related to both sensors. After the loss is not
reduced for $m=3$ epochs, we switch to hard negative mining for the remainder
of the training. The hard mining is an adaptation of the Hardnet approach of
Mishchuk et al. \cite{hardnet}. In each batch, the negative pairs of patches
are chosen by first, computing the distances between all pairs of patches
related to different modalities. Each patch is then paired with the most
similar unrelated patch from the other modality, to form the hard negative
pairs. The networks' parameters were initialized by a normal distribution,
where the asymmetric subnets were initialized identically to improve
convergence. \begin{table*}[th]
\centering
\par%
\begin{tabular}
[c]{cccccccccc}\hline
\textbf{Network/descriptor} & \textbf{Field} & \textbf{Forest} &
\textbf{Indoor} & \textbf{Mountain} & \textbf{Old building} & \textbf{Street}
& \textbf{Urban} & \textbf{Water} & \textbf{Mean}\\\hline
\multicolumn{10}{c}{\textbf{Engineered Features}}\\
\multicolumn{1}{l}{SIFT \cite{lowe2004distinctive}} & 39.44 & 11.39 & 10.13 &
28.63 & 19.69 & 31.14 & 10.95 & 40.33 & 23.95\\
\multicolumn{1}{l}{Inv SIFT \cite{MI-SIFT}} & 34.01 & 22.75 & 12.77 & 22.05 &
15.99 & 25.24 & 17.44 & 32.33 & 24.42\\
\multicolumn{1}{l}{LGHD \cite{lghd2015}} & 16.52 & 3.78 & 7.91 & 10.66 &
7.91 & 6.55 & 7.21 & 12.76 & 9.16\\
\multicolumn{1}{l}{LSS \cite{shechtman2007matching}} & 46 & 42.48 & 37.14 &
42.5 & 42.35 & 44.5 & 34.9 & 46 & 42.65\\
\multicolumn{1}{l}{DASC \cite{kim2015dasc}} & 46.68 & 35.38 & 23.19 & 41.29 &
38.07 & 39.02 & 12.28 & 45.6 & 36.68\\\hline
\multicolumn{10}{c}{\textbf{Learning-based}}\\
\multicolumn{1}{l}{Siamese \cite{Aguilera_cvprw_2016}} & 15.79 & 10.76 &
11.6 & 11.15 & 5.27 & 7.51 & 4.6 & 10.21 & 9.61\\
\multicolumn{1}{l}{Pseudo Siamese \cite{Aguilera_cvprw_2016}} & 17.01 & 9.82 &
11.17 & 11.86 & 6.75 & 8.25 & 5.65 & 12.04 & 10.32\\
\multicolumn{1}{l}{2Channe l\cite{Aguilera_cvprw_2016}} & 9.96 & \textbf{0.12}
& 4.4 & 8.89 & 2.3 & 2.18 & 1.58 & 6.4 & 4.47\\
\multicolumn{1}{l}{Q-Net 2P-4N \cite{aguilera2017cross}} & 17.01 & 2.70 &
6.16 & 9.61 & 4.61 & 3.99 & 2.83 & 8.44 & 6.86\\
\multicolumn{1}{l}{SCFDM \cite{SCFDM}} & 7.91 & 0.87 & 3.93 & 5.07 & 2.27 &
2.22 & 0.95 & 4.75 & 3.48\\
\multicolumn{1}{l}{L2-Net \cite{8100132}} & 16.77 & 0.76 & 2.07 & 5.98 &
1.89 & 2.83 & \textbf{0.62 } & 11.11 & 5.25\\
\multicolumn{1}{l}{HardNet \cite{hardnet}} & 10.89 & 0.22 & \textbf{1.87} &
3.09 & 1.32 & \textbf{1.30} & 1.19 & 2.54 & 2.80\\
\multicolumn{1}{l}{Exp-TLoss \cite{Wang_2019_ICCV}} & 5.55 & 0.24 & 2.30 &
1.51 & 1.45 & 2.15 & 1.44 & 1.95 & 2.07\\\hline
\multicolumn{10}{c}{\textbf{Hybrid CNN}}\\
\multicolumn{1}{l}{Hybrid-CL} & 4.61 & 0.22 & 2.52 & 2.69 & 1.52 & 1.34 &
1.55 & 2.22 & 1.93\\
\multicolumn{1}{l}{D-Hybrid-CL} & \textbf{4.4} & 0.20 & 2.48 & \textbf{1.50} &
\textbf{1.19} & 1.93 & 0.78 & \textbf{1.56} & \textbf{1.7}\\\hline
\end{tabular}
\caption{\textbf{Patch matching results evaluated using the VIS-NIR dataset.}
The patches were extracted as in Aguilera et al.
\cite{Aguilera_cvprw_2016,aguilera2017cross}. The accuracy is given in terms
of the FPR95 score. Hybrid-CL is the implantation of the proposed scheme using
the CNN, as in Section \ref{subsec:CNN architecture}, while D-Hybrid-CL
utilizes the deeper CNN, as in Wang et al. {\cite{Wang_2019_ICCV}.}}%
\label{tab:VIS-NIR_keypoint_res}%
\end{table*}

\subsection{VIS-NIR benchmark}

\label{sec:Vis-Nir benchmark}

The proposed Hybrid scheme was experimentally evaluated using the VIS-NIR
dataset \cite{Aguilera_cvprw_2016}\footnote{https://github.com/ngunsu/lcsis},
and was compared to the previous results, all using the same experimental
setup as in Aguilera et al. \cite{Aguilera_cvprw_2016,aguilera2017cross} and
En et al. \cite{en2018ts}. All of the schemes were trained using the 'Country'
category, where we utilized $80\%$ of the given training pairs of patches for
training, and the remaining $20\%$ for evaluation. The results are reported in
Table \ref{tab:VIS-NIR_keypoint_res}. We compared against several classes of
previous works. First, we studied the accuracy of handcrafted local
descriptors such as LSS \cite{shechtman2007matching}, DASC \cite{kim2015dasc},
LGHD \cite{lghd2015}, SIFT \cite{lowe2004distinctive} and MI-SIFT
\cite{MI-SIFT} using their publicly available code. Such schemes are shown to
be notably outperformed by recent learning-based approaches, including ours.
The LSS and DASC performed worse, as such descriptors aim to encode
corresponding large-scale geometrical patterns that might not exist in
multisensor images. Second, we compared to HardNet \cite{hardnet} and L2-Net
{\ \cite{8100132}} that are recent, general-purpose patch matching schemes,
based an a siamese CNN, which were specifically trained using the VIS-NIR
dataset protocol by Wang et al. {\ \cite{Wang_2019_ICCV}}. Last, we compared
to recent the SOTA multi modal results of SCFDM \cite{SCFDM} and Wang et al.
{\ \cite{Wang_2019_ICCV}, }that also utilized the same dataset and
experimental se up.

Our scheme outperforms HardNet {\ \cite{hardnet} }and L2-Net {\ \cite{8100132}%
} by 64\% and 200\%, respectively. Wang et al. {\ \cite{Wang_2019_ICCV}
}presented the recent SOTA results, using a deeper CNN than the one detailed
in Section \ref{subsec:CNN architecture}. Hence, we report the results of
\textit{two} implementations of the proposed Hybrid scheme. The first
Hybrid-CL (Contrastive Loss), utilizes the CNN detailed in Table
\ref{tab:L2BranchArchitecture} as a base CNN for the symmetric and asymmetric
branches, and is further used in the following sections. The second, denoted
D-Hybrid-CL was implemented using the deeper CNN of Wang et al.
{\ \cite{Wang_2019_ICCV}}. Both of our schemes outperform the previous SOTA
(Wang et al. {\ \cite{Wang_2019_ICCV}}) \ by 17\%, where the deeper\ CNN is
shown to provide additional accuracy.

\subsection{ En et al. \cite{en2018ts} benchmark}

\label{sec:ani}

We also evaluated the proposed scheme using the experimental setup proposed by
En et al. \cite{en2018ts}\footnote{https://github.com/ensv/TS-Net} where the
VEDAI \cite{razakarivony2016vehicle}, CUHK \cite{wang2009face} and VIS-NIR
\cite{Aguilera_cvprw_2016} datasets were sampled on a uniform grid, and the
results are reported in Table \ref{tab:grid_res}. We quote the results
reported by En et al. \cite{en2018ts} for these datasets and setup, and
trained the publicly available code of Aguilera et al.
\cite{Aguilera_cvprw_2016,aguilera2017cross} and the proposed Hybrid scheme,
using $70\%$ of the data in each dataset for training, $10\%$ for validation
and $20\%$ for testing. As before we compared with the results of the SIFT
\cite{lowe2004distinctive} and modality-invariant descriptors: LSS
\cite{shechtman2007matching}, DASC \cite{kim2015dasc}, LGHD \cite{lghd2015}
and MI-SIFT \cite{MI-SIFT}.

It follows that the proposed approach significantly outperformed the previous
schemes\ for the UVIS-NIR and CUHK datasets yielding an average error that is
threefold more accurate. For the VEDAI dataset, both Aguilera et al.
\cite{Aguilera_cvprw_2016}, and the proposed scheme achieved a zero error. To
further validate these results, we trained the VEDAI matching network 10
times, each time starting from a different random initialization of the CNN
weights. We achieved zero error in all of the trained CNNs and a corresponding
standard deviation (STD) of $\sigma=0$. We also computed the FPR99 and also
achieved zero error and STD of $\sigma=0$. This superior performance is
achieved without applying hard-mining, emphasizing that the Hybrid CNN
formulation is the one doing the heavy lifting. In particular, comparing to
the TS-Net \cite{en2018ts}, our approach is \textit{3-20 times more}\textbf{
}accurate. \begin{table}[tbh]
\centering
\par%
\begin{tabular}
[c]{cccc}\hline
Network/descriptor & VEDAI & CUHK & UVIS-NIR\\\hline
\multicolumn{4}{c}{Engineered Features}\\
\multicolumn{1}{l}{SIFT\cite{lowe2004distinctive}} & 42.74 & 5.87 & 32.53\\
\multicolumn{1}{l}{Inv SIFT \cite{MI-SIFT}} & 11.33 & 7.34 & 27.71\\
\multicolumn{1}{l}{LSS\cite{shechtman2007matching}} & 39.9 & 43.11 & 42.25\\
\multicolumn{1}{l}{DASC\cite{kim2015dasc}} & 8.9 & 43.05 & 38.1\\
\multicolumn{1}{l}{LGHD\cite{lghd2015}} & 1.31 & 0.65 & 10.76\\\hline
\multicolumn{4}{c}{Learning-based}\\
\multicolumn{1}{l}{2Channel\cite{Aguilera_cvprw_2016}} & \textbf{0} & 0.39 &
11.32\\
\multicolumn{1}{l}{Q-Net 2P-4N\cite{aguilera2017cross}} & 0.78 & 0.9 & 22.5\\
\multicolumn{1}{l}{Siamese\cite{en2018ts}} & 0.84 & 3.38 & 13.17\\
\multicolumn{1}{l}{Pseudo Siamese\cite{en2018ts}} & 1.37 & 3.7 & 15.6\\
\multicolumn{1}{l}{TS-Net\cite{en2018ts}} & 0.45 & 2.77 & 11.86\\\hline
\multicolumn{4}{c}{Hybrid CNN}\\
\multicolumn{1}{l}{\textbf{Hybrid-CE}} & \textbf{0} & \textbf{0.05} & 3.66\\
\multicolumn{1}{l}{\textbf{Hybrid-CL}} & \textbf{0,}$\sigma=0$ & 0.1 &
\textbf{3.41}\\
\multicolumn{1}{l}{\textbf{Hybrid-CL FPR99}} & \textbf{0,}$\sigma=0$ & - &
-\\\hline
\end{tabular}
\caption{\textbf{Patch matching results evaluated using the UVIS-NIR, VEDAI
and CUHK datasets.} The patches were extracted using a uniform lattice layout
as in En et al. \cite{en2018ts}. The accuracy is given in terms of the FPR95
score. The schemes names in bold are variations of the proposed scheme, CE and
CL relates to using the Cross-Entropy and Contrastive losses, respectively.
For the VEDAI dataset, we trained the Hybrid-CL ten times, starting from
random weights and also report the standard deviation $\sigma$, and the FPR99
results.}%
\label{tab:grid_res}%
\end{table}

\subsection{Features detection}

\label{subsec:Features detection} \begin{figure*}[tbh]
\centering\subfigure[]{\includegraphics[width=0.33\textwidth]{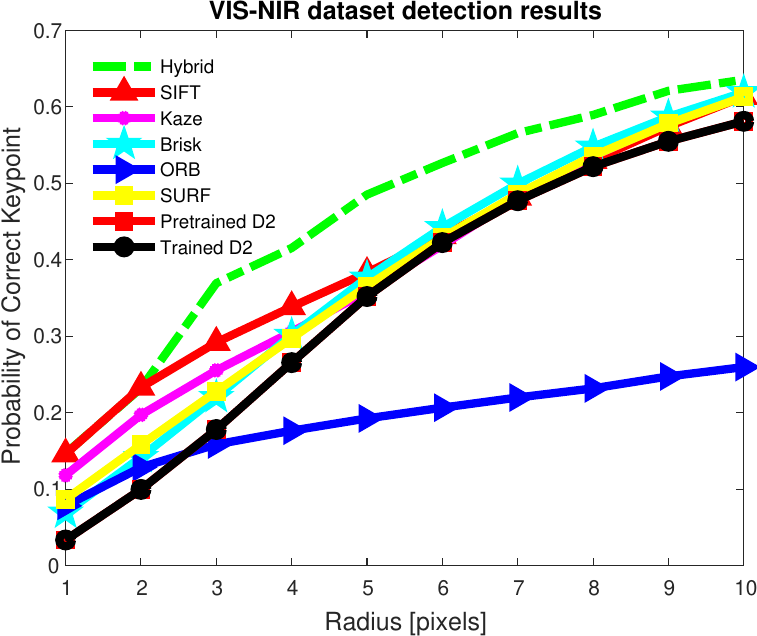}}\subfigure[]{\includegraphics[width=0.33\textwidth]{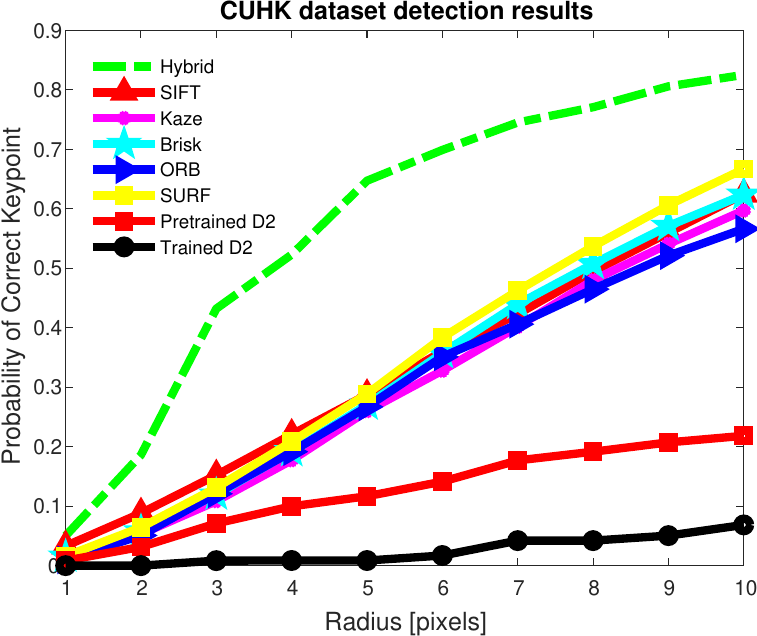}}\subfigure[]{\includegraphics[width=0.33\textwidth]{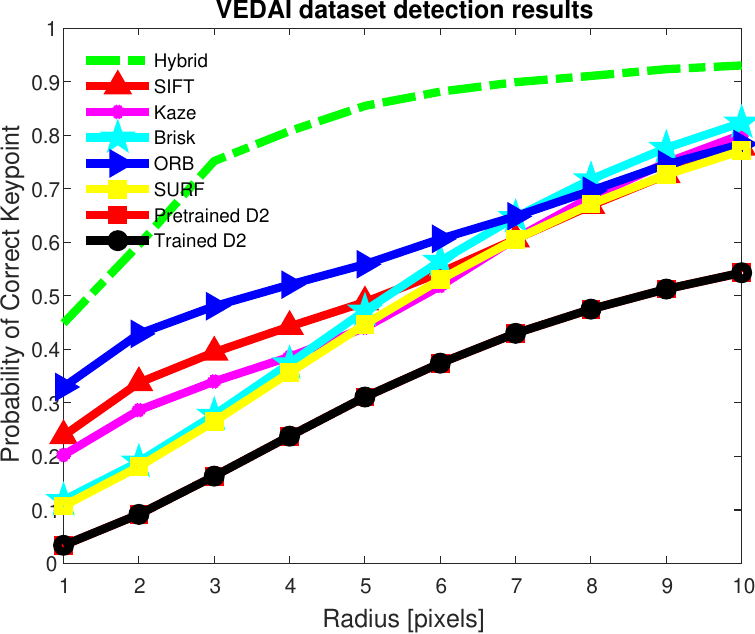}}\caption{\textbf{Feature
detection results.} We report the Probability of Correct Keypoint
\cite{choy2016universal} and compare with handcrafted detectors and the D2-Net
\cite{D2-Net}. (a) VIS-NIR dataset results. The images and detection maps are
680$\times$1024 and 78$\times$121, respectively. (b) CUHK dataset results. The
images and detection maps are 250$\times$200 and 24$\times$18, respectively.
(c) VEDAI dataset results. The images and detection maps are 512$\times$512
and 57$\times$57, respectively.}%
\label{fig:detection}%
\end{figure*}

The proposed feature point detection scheme, introduced in Section
\ref{subsec:detection}, was experimentally evaluated following Mikolajczyk and
Schmid \cite{1498756}. The detection repeatability was measured using the
VEDAI \cite{razakarivony2016vehicle}, CUHK \cite{wang2009face} and VIS-NIR
\cite{Aguilera_cvprw_2016} image sets consisting of \textit{aligned}
multimodal images. Thus, for each point $\mathbf{v}_{1}^{i}\in\mathbf{I}%
_{\mathbf{1}}$, there is a \textit{groundtruth }corresponding point
$\mathbf{v}_{2}^{i}\in\mathbf{I}_{\mathbf{2}}$. Let $\mathbf{v}_{1}^{d}%
\in\mathbf{I}_{\mathbf{1}}$ be a \textit{detected} feature point, whose
corresponding \textit{groundtruth} point is $\mathbf{v}_{2}^{d}\in
\mathbf{I}_{\mathbf{2}}$, and let $\mathbf{v}_{2}^{d^{\ast}}\in\mathbf{I}%
_{\mathbf{2}}$ be the \textit{detected} point that is the \textit{closest} to
$\mathbf{v}_{2}^{d}$. We report the Probability of Correct Keypoints
\cite{choy2016universal}%
\begin{equation}
P_{C}=P\left(  \left\vert \mathbf{v}_{2}^{d}-\mathbf{v}_{2}^{d^{\ast}%
}\right\vert _{2}\leq r\right)  ,\text{ }r\in%
\mathbb{R}
^{+}, \label{equ:detection}%
\end{equation}
that is the average probability (over all detected points $\mathbf{v}_{1}%
^{d}\in\mathbf{I}_{\mathbf{1}}$) that the closest feature point $\mathbf{v}%
_{2}^{d}\in\mathbf{I}_{\mathbf{2}}$ was detected within a radius $r$ of the
groundtruth match $\mathbf{v}_{2}^{d}$. For each pair of images, we compute
$P_{C}$ twice by switching $\mathbf{I}_{\mathbf{1}}$ and $\mathbf{I}%
_{\mathbf{2}}$ and averaging the results.

The proposed Hybrid detector provides a fixed detection grid, while detectors
such as SIFT detect a varying number of feature points per image, as they
utilize a cornerness score that is compared with a detection threshold. For
instance, the VIS-NIR dataset consists of 680$\times$1024 images, resulting in
78$\times$121 detection maps, while the SIFT typically detects $\sim500$
feature points at most per image. Denote $N_{s}\left(  I\right)  $ as the
number of SIFT-based feature points detected in an image $I$. A denser
detection grid might result in a higher detection probability $P_{C}$. Thus,
to allow a fair comparison, we utilize $N_{s}\left(  I\right)  $ feature
points per image for \textit{all} detectors. The points in the Hybrid-based
detection map are sorted by their activation, and the leading $N_{s}\left(
I\right)  $ feature points are retained.

We first compared the proposed detector to classical handcrafted detectors:
SIFT \cite{lowe2004distinctive}, SURF\ \cite{bay2008speeded},
KAZE\ \cite{KAZE}, BRISK\ \cite{BRISK}, and ORB \cite{ORB} detectors.
Handcrafted detectors, in contrast to descriptors are not modality-specific,
making the comparison valid. We also compared to the deep learning-based
D2-Net \cite{D2-Net} \ detection and matching CNN. For that, we first compare
to a pretrained D2-Net, that was trained on RGB images with significant
appearance variations (day, night, shadows etc.). We also trained a D2-Net on
the different multimodal datasets according to their experimental protocols.

The detection results are shown in Fig. \ref{fig:detection} where we report
the cumulative detection probabilities as in Eq. \ref{equ:detection}. It
follows that the proposed Hybrid detector significantly outperformed the
handcrafted detectors and the pretrained D2-Net. The D2-Net performed best for
the VIS-NIR dataset, that consists of scenes that are most similar to the
D2-Net dataset. The trained D2-Net performed worse than the pretrained D2-Net
and we attribute that to the smaller training sets. The pretrained D2-Net CNN
was trained using the MegaDepth dataset \cite{MegaDepthLi18} consisting of
102,681 images, that is a multiple orders of magnitude larger than the VEDAI,
CUHK and VIS-NIR datasets.

We show qualitative \textit{detection results} in Figs.
\ref{fig:qualitative detect1}, \ref{fig:qualitative detect2} and in the
Supplementary materials. We compare the detections of the proposed Hybrid
scheme, SIFT and the pretrained D2-Net. For each approach and particular image
modality (either VIS or NIR) we show the 200 detections having the largest
detection scores, that were classified as either inliers or outliers. For each
detected feature point $x$ (say in Fig. \ref{fig:qualitative detect1}a) we
search for a detected point in the other modality image (Fig.
\ref{fig:qualitative detect1}b) that is within $r=5$ pixels, same as in Eq.
\ref{equ:detection}. If such a point exists, $x$ is considered an inlier
(green), and an outlier (red) otherwise. To exemplify the general
applicability of the proposed detection scheme, we chose an image showing a
natural scene in Fig. \ref{fig:qualitative detect1} mainly consisting of
textures, in contrast to the urban environment in Fig.
\ref{fig:qualitative detect2}, that is characterized by significant edges. The
Hybrid approach detects elongated edge-like features that are known to be
salient and \textit{repeatable} to appearance variations. It outperforms both
other schemes in terms of repeatability and the number of inliers, although
the D2-Net \cite{D2-Net} was mainly trained on the urban views in the
MegaDepth dataset \cite{MegaDepthLi18}. \begin{figure}[tbh]
\centering%
\begin{tabular}
[c]{ll}%
\subfigure[]{\includegraphics[width=0.35\linewidth]{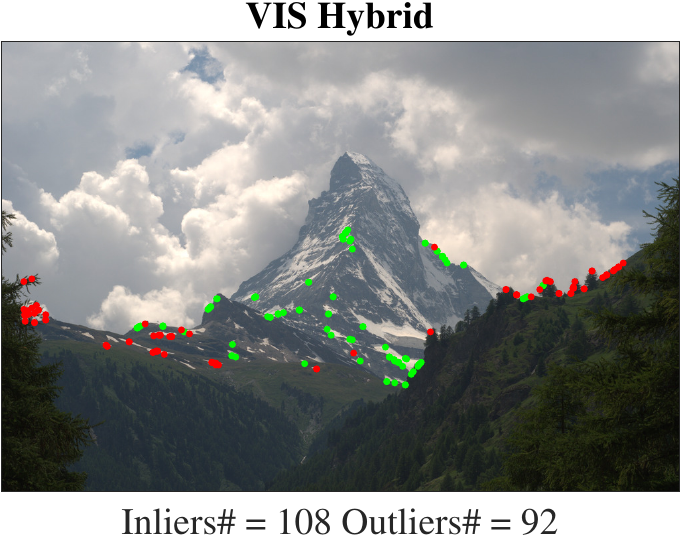}} &
\subfigure[]{\includegraphics[width=0.35\linewidth]{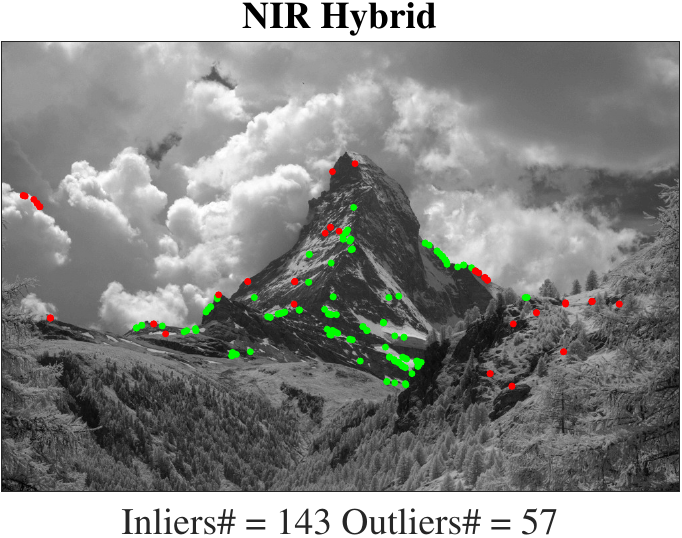}}\\
\subfigure[]{\includegraphics[width=0.35\linewidth]{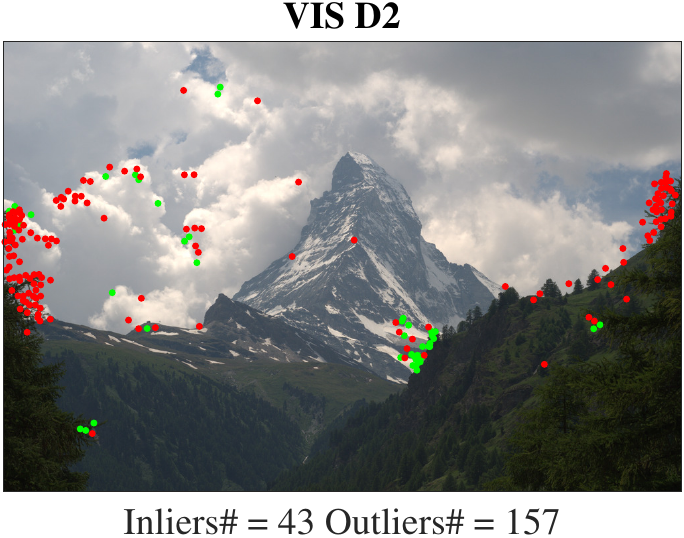}} &
\subfigure[]{\includegraphics[width=0.35\linewidth]{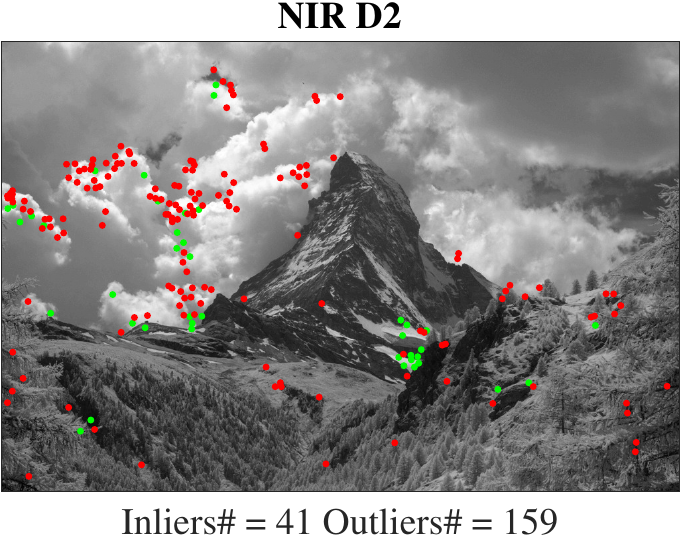}}\\
\subfigure[]{\includegraphics[width=0.35\linewidth]{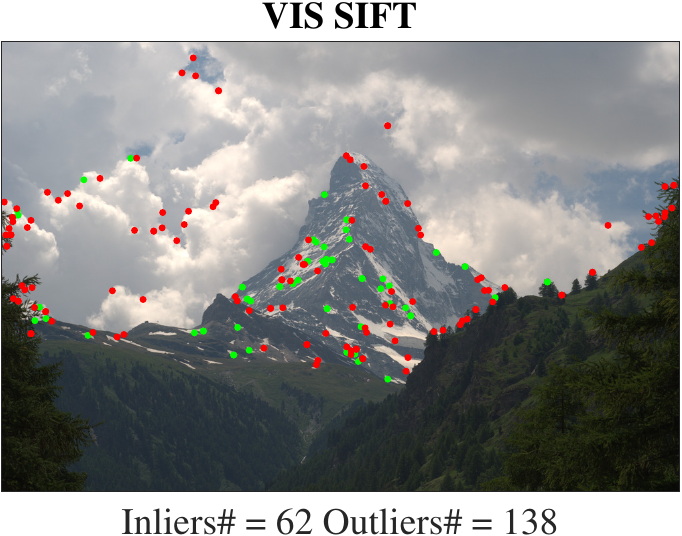}} &
\subfigure[]{\includegraphics[width=0.35\linewidth]{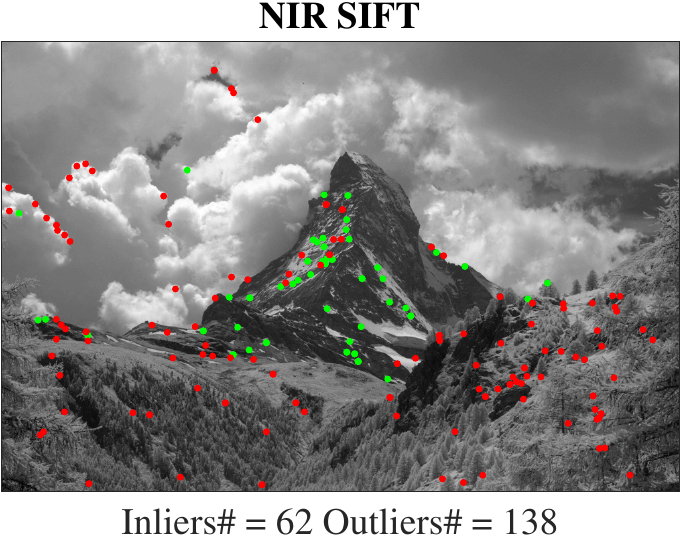}}
\end{tabular}
\caption{\textbf{Qualitative feature detection results.} For each approach we
show the 200 detections having the largest detection scores. The
\textcolor{green}{inliers} are marked in green, while the
\textcolor{red}{outliers} are marked in red. Please zoom-in to view the
detected points.}%
\label{fig:qualitative detect1}%
\end{figure}\begin{figure}[tbh]
\centering%
\begin{tabular}
[c]{ll}%
\subfigure[]{\includegraphics[width=0.35\linewidth]{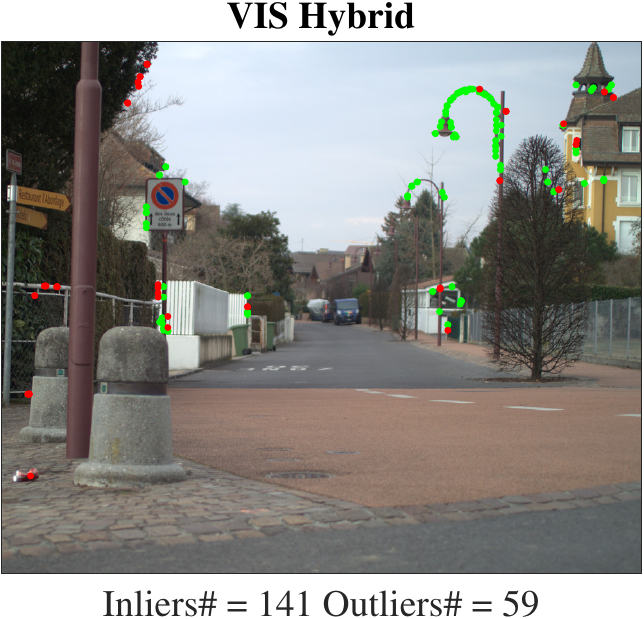}} &
\subfigure[]{\includegraphics[width=0.35\linewidth]{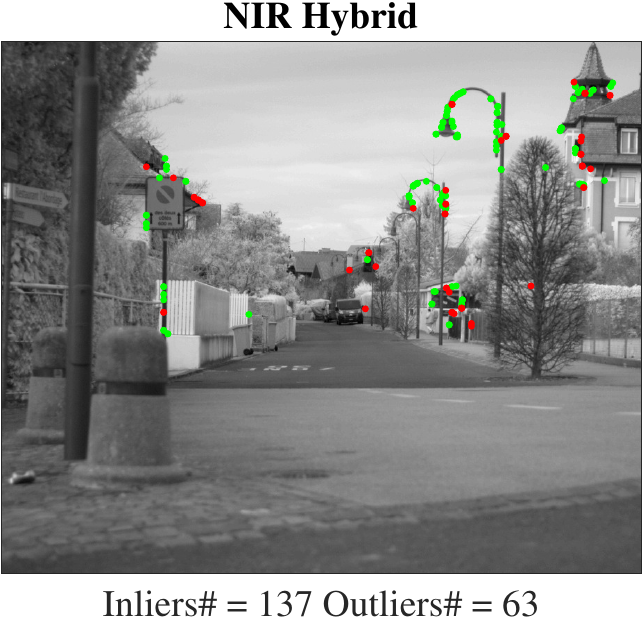}}\\
\subfigure[]{\includegraphics[width=0.35\linewidth]{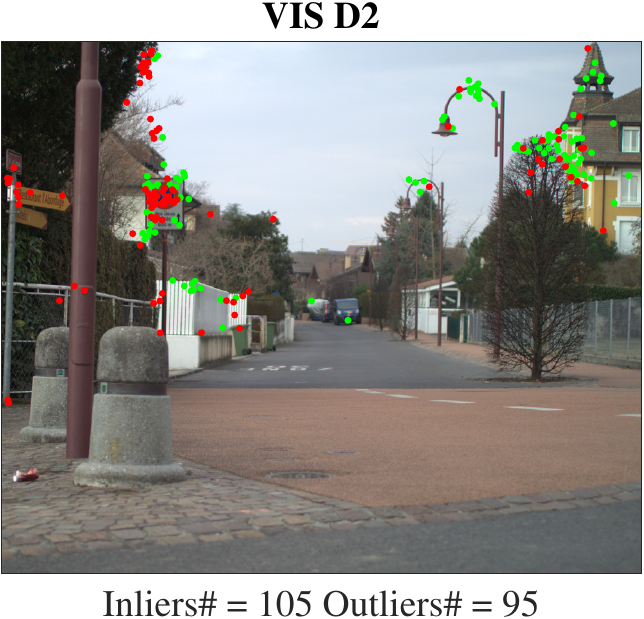}} &
\subfigure[]{\includegraphics[width=0.35\linewidth]{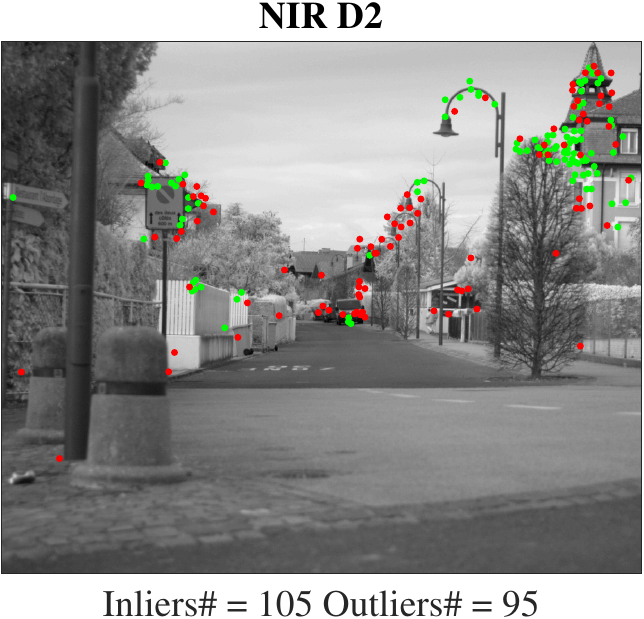}}\\
\subfigure[]{\includegraphics[width=0.35\linewidth]{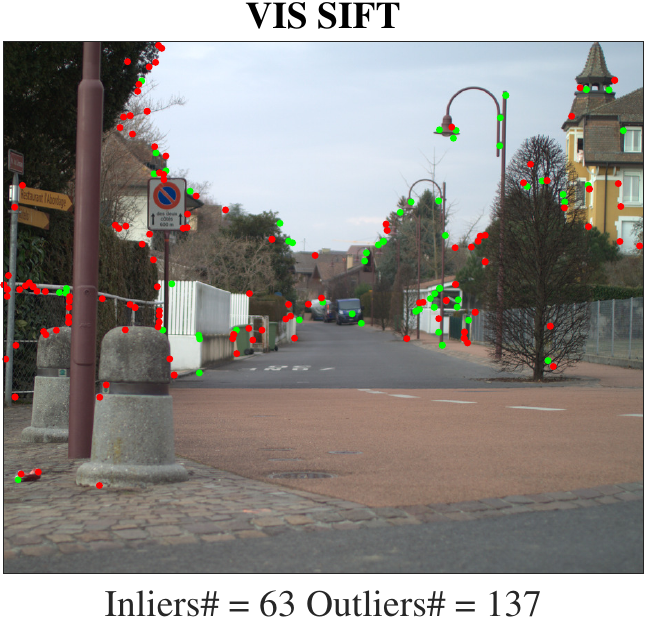}} &
\subfigure[]{\includegraphics[width=0.35\linewidth]{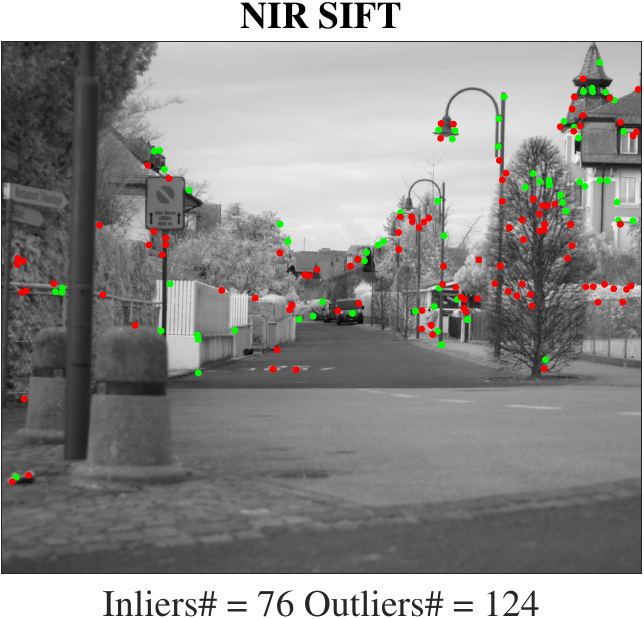}}
\end{tabular}
\caption{\textbf{Qualitative feature detection results.} For each approach we
show the 200 detections having the largest detection scores. The
\textcolor{green}{inliers} are marked in green, while the
\textcolor{red}{outliers} are marked in red. Please zoom-in to view the
detected points.}%
\label{fig:qualitative detect2}%
\end{figure}

\begin{figure}[tbh]
\centering%
\begin{tabular}
[c]{l}%
\subfigure[]{\includegraphics[width=0.95\linewidth]{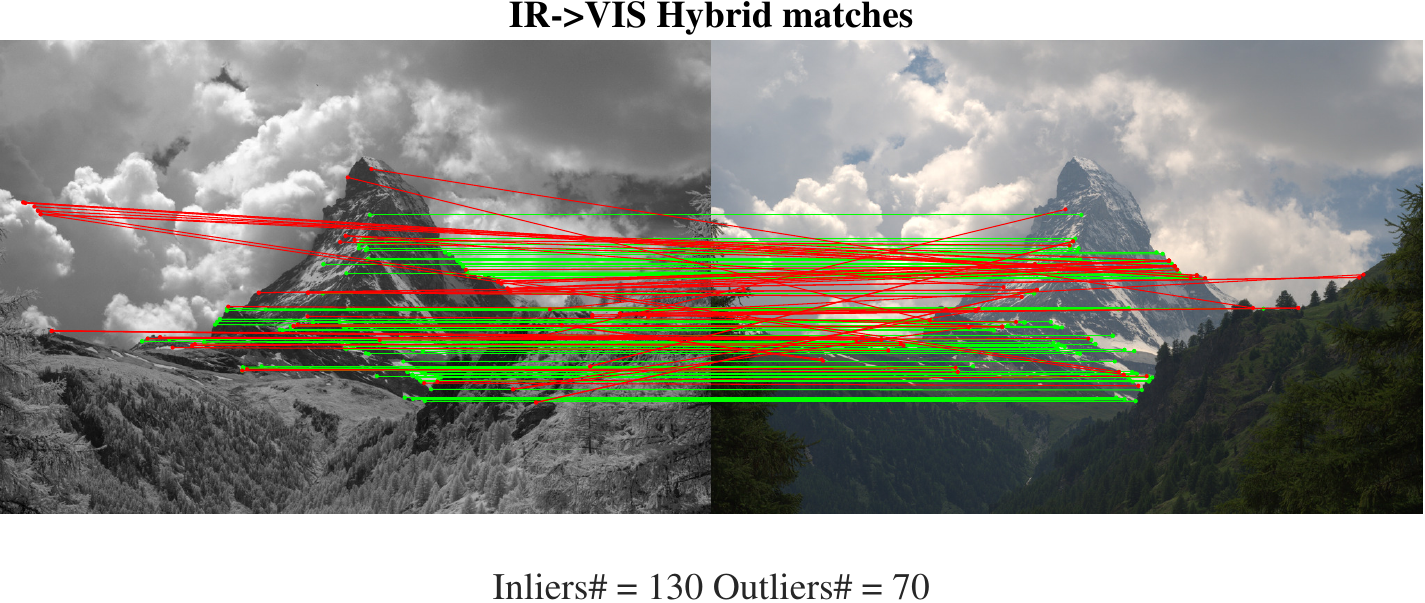}}\\
\subfigure[]{\includegraphics[width=0.95\linewidth]{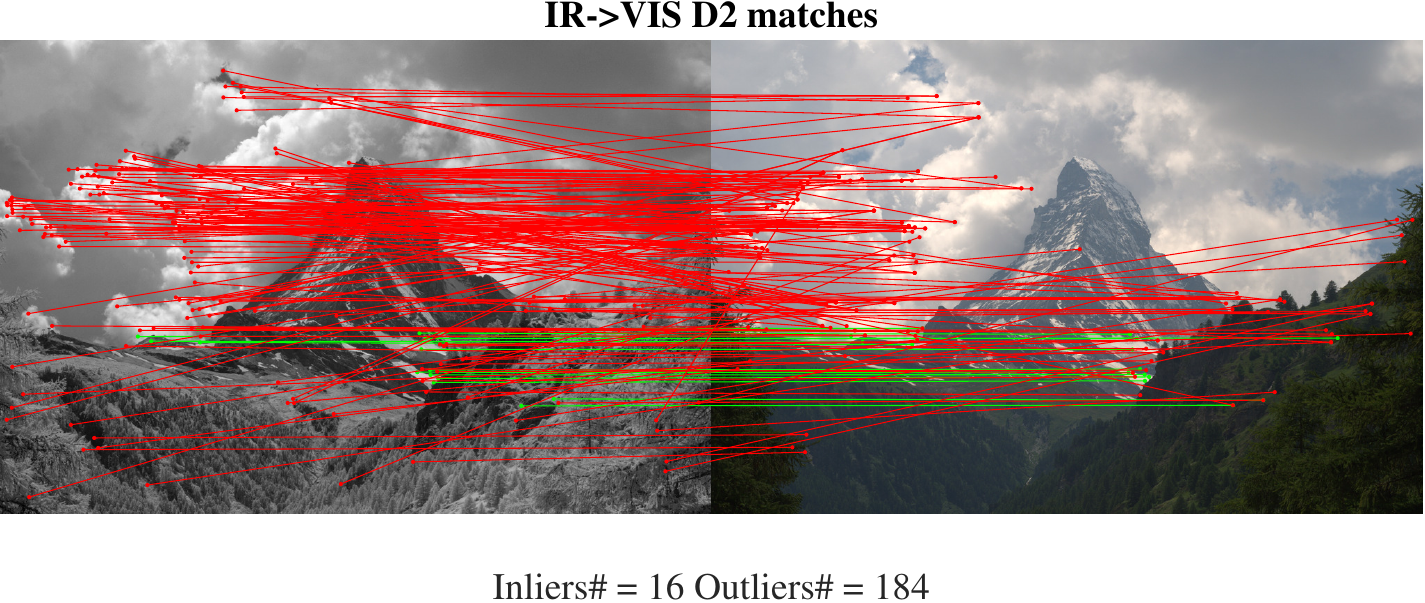}}\\
\subfigure[]{\includegraphics[width=0.95\linewidth]{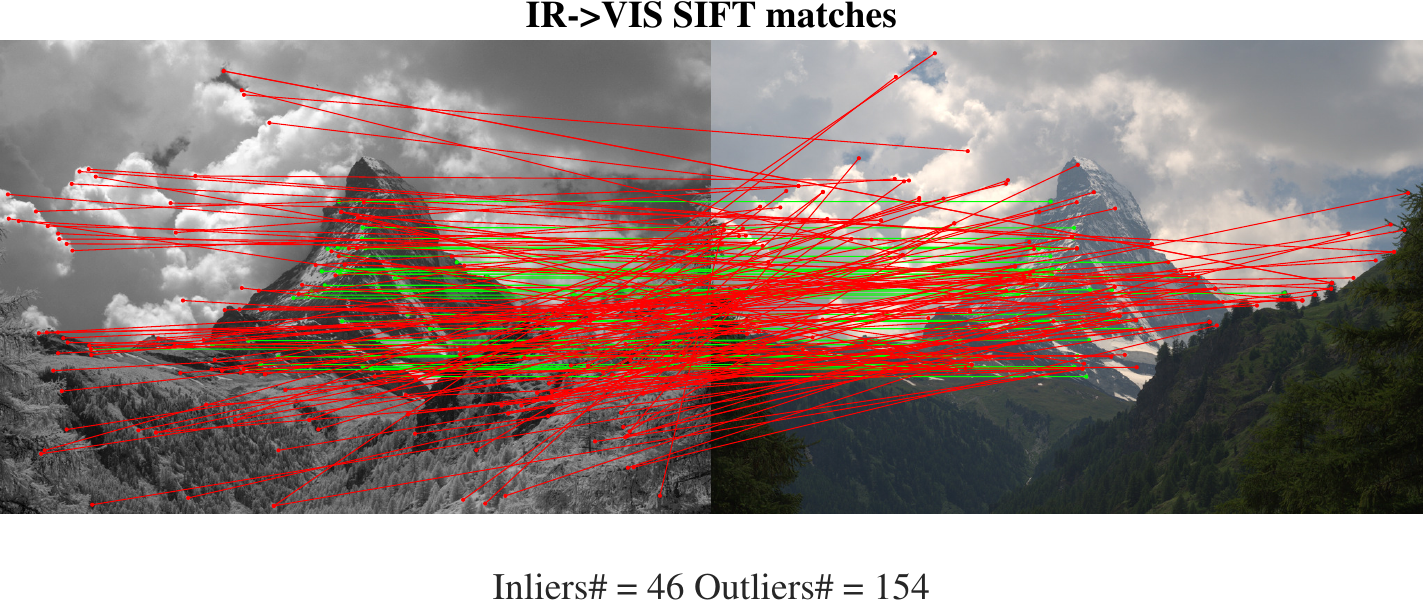}}
\end{tabular}
\caption{\textbf{Qualitative feature detection and matching results.} For each
approach we match the top 200 detections in the IR image to top 400 detections
in the VIS image, based on the $L_{2}$ distance of their descriptors. The
\textcolor{green}{inliers} that are correctly matched within a $r=5$ are
marked in green, while the \textcolor{red}{outliers} are marked in red.}%
\label{fig:detect-match1}%
\end{figure}

\begin{figure}[tbh]
\centering%
\begin{tabular}
[c]{l}%
\subfigure[]{\includegraphics[width=0.95\linewidth]{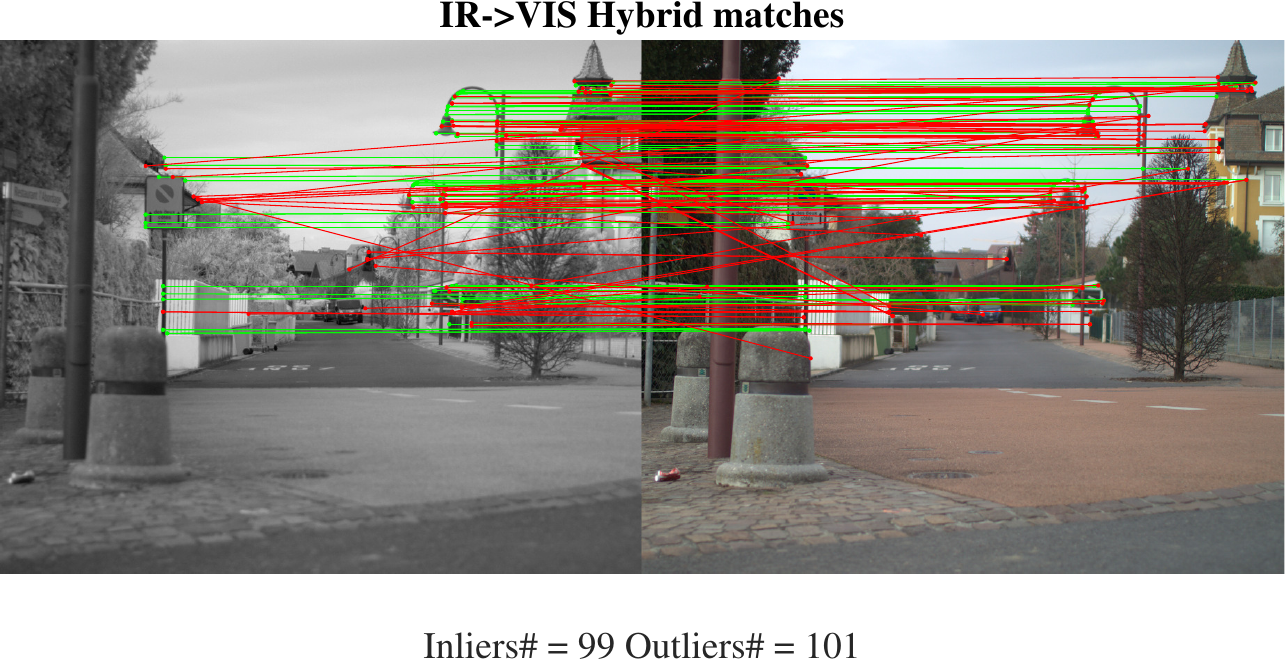}}\\
\subfigure[]{\includegraphics[width=0.95\linewidth]{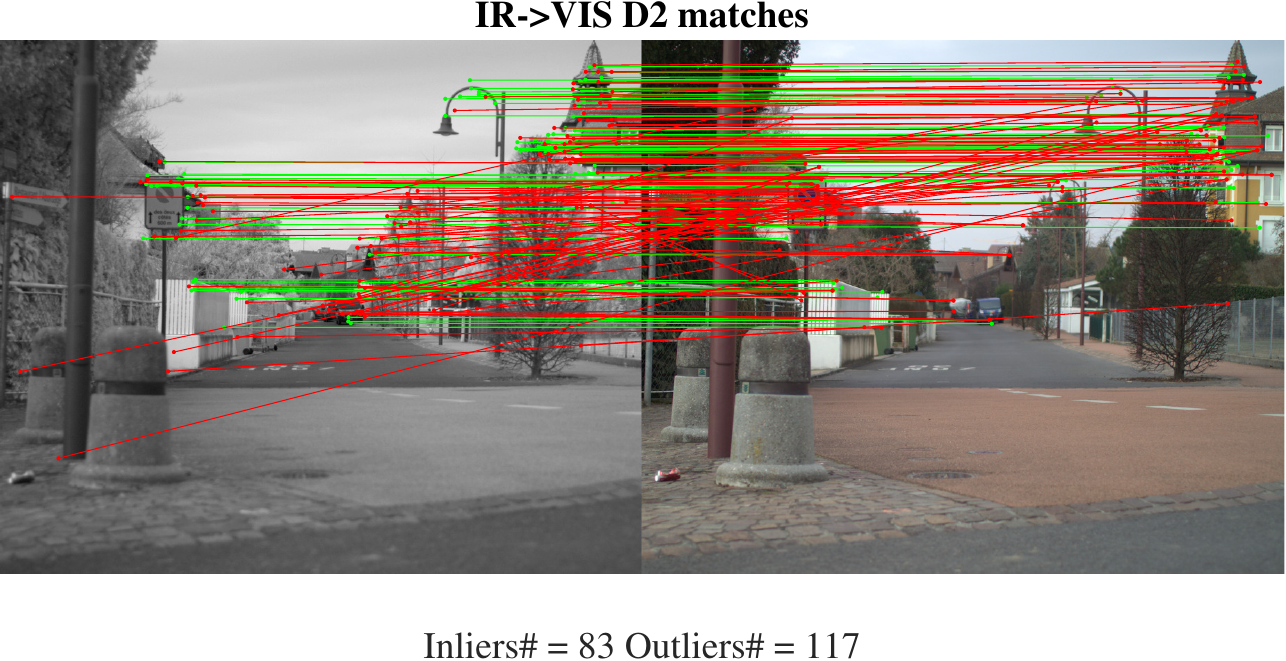}}\\
\subfigure[]{\includegraphics[width=0.95\linewidth]{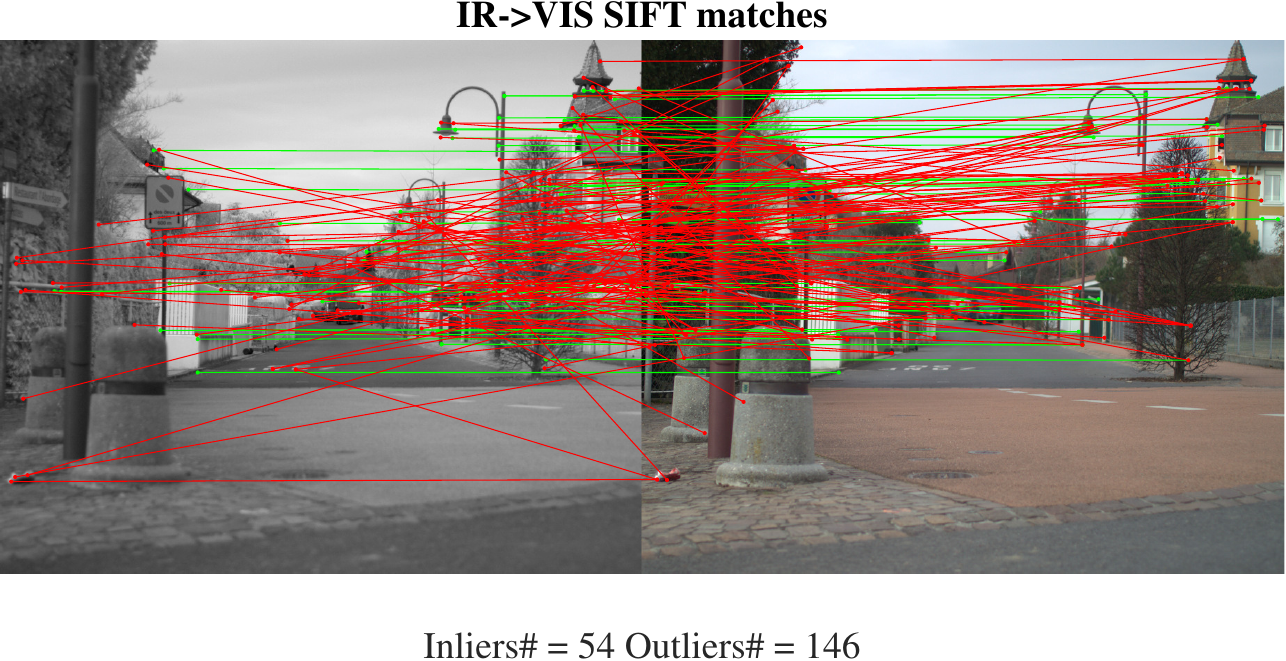}}
\end{tabular}
\caption{\textbf{Qualitative feature detection and matching results.} For each
approach we match the top 200 detections in the IR image to top 400 detections
in the VIS image, based on the $L_{2}$ distance of their descriptors. The
\textcolor{green}{inliers} that are correctly matched within a $r=5$ are
marked in green, while the \textcolor{red}{outliers} are marked in red.}%
\label{fig:detect-match2}%
\end{figure}

We also tested \textit{combined} \textit{detection and matching}, where the
feature points were first detected and then matched by the same schemes as in
Figs. \ref{fig:qualitative detect1}-\ref{fig:qualitative detect2} to evaluate
their overall accuracy. The results are shown in Figs. \ref{fig:detect-match1}
and \ref{fig:detect-match2}, where we first detected 200 feature points in the
IR image and matched them to 400 points detected in the corresponding RGB
image, based on the $L_{2}$ distance between the descriptors. As the NIR and
RGB images are pixelwise aligned, we considered the matches as inliers when
the spatial distance between $L_{2}$-matched points was less than $r=5$
pixels. Otherwise the detections were considered outliers. The proposed Hybrid
scheme outperformed the other schemes in \textit{all} cases in terms of the
number of inliers. In particular, the upside is significant for the natural
scene in Fig. \ref{fig:detect-match1}, compared to the urban scene in Fig.
\ref{fig:detect-match2}. We attribute that to D2 being trained using the urban
views in the MegaDepth dataset \cite{MegaDepthLi18}, while the SIFT is
handcrafted to identify corners using the DoG detector, that is more
applicable to urban scenes. We also tested the complementary matchings of
VIS-to-IR, and achieved similar results.

\subsection{Ablation study}

\label{sec:Auxiliary}

We conducted an ablation study by comparing the results of the proposed Hybrid
CNN to CNN formulations where one of the algorithmic components is omitted or
changed. This allows to evaluate the contribution of each of the proposed
components. Thus, we first compare to using only the Siamese and Asymmetric
CNNs and also compare the results of training the Hybrid CNN using the
multi-loss approach as in Section \ref{sec:Proposed Model}, to using a single
loss ($L_{H}$ in Fig. \ref{fig:hybridModelIllustration}) while omitting the
auxiliary losses $L_{S}$ and $L_{A}$. We also show the added value of applying
hard negative mining. The different CNNs were applied to the VIS-NIR
\cite{Aguilera_cvprw_2016} dataset using the experimental setup by En et al.
\cite{en2018ts}, the same as in Section \ref{sec:ani}. The results are
reported in Table \ref{table:ablation}, where the Hybrid CNN is shown to
outperform the Siamese and asymmetric CNNs when using either the CE or CL
losses. The use of hardmining improves the accuracy significantly, for the
CL-based results, in particular, by close to 6.5\%. The auxiliary losses in
the Hybrid scheme provides an additional accuracy improvement of 1.1\%.
\begin{table}[tbh]
\centering
\par%
\begin{tabular}
[c]{ccccc}\hline
Network & Loss & \#Losses & Hard Mining & VIS-NIR\\\hline\hline
\multicolumn{1}{l}{Hybrid} & CE & 1 & + & 4.72\\
\multicolumn{1}{l}{Symmetric} & CE & 1 & + & 4.2\\
\multicolumn{1}{l}{Asymmetric} & CE & 1 & + & 5.8\\
\multicolumn{1}{l}{Hybrid} & CE & 3 & - & 5.78\\
\multicolumn{1}{l}{Hybrid} & CE & 3 & + & 3.66\\\hline
\multicolumn{1}{l}{Hybrid} & CL & 1 & + & 4.52\\
\multicolumn{1}{l}{Symmetric} & CL & 1 & + & 9.56\\
\multicolumn{1}{l}{Asymmetric} & CL & 1 & + & 4.32\\
\multicolumn{1}{l}{Hybrid} & CL & 3 & - & 9.90\\
\multicolumn{1}{l}{Hybrid} & CL & 3 & + & 3.41\\\hline
\end{tabular}
\caption{Ablation results evaluated using the VIS-NIR dataset, where the
patches were extracted using a uniform lattice layout as in En et al.
\cite{en2018ts}. The accuracy is given in terms of the FPR95 score. CE and CL
relates to applying the Cross-Entropy and Contrastive losses, respectively.
\#Losses relates to the number of losses used for training. A single loss
relates to training the CNN using a single loss $L_{H}$, as in Fig.
\ref{fig:hybridModelIllustration}.}%
\label{table:ablation}%
\end{table}

\section{Conclusions}

\label{sec:Conclusions}

In this work, we presented a Deep-Learning approach for the detection and
matching of feature points in multimodal images, that utilizes a novel Hybrid
CNN formulation consisting of two CNN sub-networks. The first is a Siamese
(weight-sharing) CNN, while the second is an asymmetric (non-weigh-sharing)
CNN. A novel feature points detection approach is derived by backtracking
through the Siamese subnetwork activations, following the dominant
activations. We show that the matching accuracy is improved by applying
multi-loss learning to the Siamese and asymmetric sub-networks, alongside the
principal output loss. The proposed scheme is experimentally shown to
outperform state-of-the-art approaches when applied to multiple multimodal
image datasets. It significantly reduces the matchings errors by two to
threefold and outperforms state-of-the-art detectors such as SIFT, SURF and
the D2-net in terms of detection repeatability.

\bibliographystyle{plain}
\bibliography{bibliographyList}

\end{document}